\documentclass[sigconf,nonacm]{acmart}

\makeatletter
\@ACM@balancefalse
\makeatother

\usepackage{booktabs}
\PassOptionsToPackage{expansion=false}{microtype}
\usepackage{url}
\usepackage{listings}
\usepackage{graphicx}
\usepackage{multirow}
\usepackage{longtable}
\usepackage{array}
\newcolumntype{L}[1]{>{\raggedright\arraybackslash}p{#1}}
\usepackage{calc}
\usepackage{tikz}
\usetikzlibrary{arrows.meta,positioning,fit,calc}

\AtBeginDocument{%
  }

\providecommand{\tightlist}{%
  \setlength{\itemsep}{0pt}\setlength{\parskip}{0pt}}
\providecommand{\passthrough}[1]{#1}
\setcopyright{none}

\begin{document}

\title{Evaluating Governed {LLM}-Driven {ML} Experimentation: Lifecycle, Provenance, and Failure Metrics}
\titlenote{Presented at the KDD 2026 Workshop on Evaluation and
  Trustworthiness of Agentic AI, Jeju, Korea, 9 August 2026. The workshop is
  non-archival: its papers are posted on the workshop website and are not part
  of the KDD proceedings. This version corrects the workshop version: it
  rewords several claims, reports the starting point of the L40S
  replication, corrects and adds references, updates the
  artifact-availability statement, and includes the supplement as
  appendices.}

\author{Dowling Wong}
\email{dowling.wong@kit.edu}
\affiliation{%
  \institution{Karlsruhe Institute of Technology}
  \city{Karlsruhe}
  \country{Germany}}
\renewcommand{\shortauthors}{Dowling Wong}

\begin{abstract}
Autonomous agents that run ML experiments are usually judged by the best score
they reach. That score says nothing about how it was reached: whether the agent
stayed within the files it was allowed to edit, whether failed trials were kept
or silently dropped, and whether each keep or discard decision can be traced to
its evidence. We propose evaluating such systems on their experimental process
as well as their outcome, and we build a governed harness that makes this
possible. A control plane, not the LLM that proposes changes, owns each trial's
lifecycle, checks its edit scope, decides keep or discard, and appends one
record per trial to an append-only ledger. Governance metrics such as the
invalid-action rate, the repeated-bad rate and provenance completeness are
computed from that ledger. Across six ML tasks, from synthetic logistic
regression to language-model training, and 1{,}445 trials, the same contract
applied unchanged. The ledger exposed failures that a task score hides: a
no-memory arm in which all 120 trials failed, a benchmark adapter whose bounded
proposals ran out, and a reset failure that reproducibility hashes revealed,
after which the affected campaigns were rerun. In a matched counterfactual, a
loop that logs only successful trials left no record of 43 of 50 trials, while
the governed ledger recorded all 50. Memory ablations show that memory's effect on repeated bad proposals depends
on the task. We release the ledgers of all 1{,}445 trials, so that every number
reported for them can be recomputed.
\end{abstract}



\maketitle
\ifdefined\pdfinfo\pdfinfo{/Author (Dowling Wong)}\fi

\section{Introduction}\label{introduction}
\begin{figure*}[t]
\centering
\includegraphics[width=0.78\textwidth]{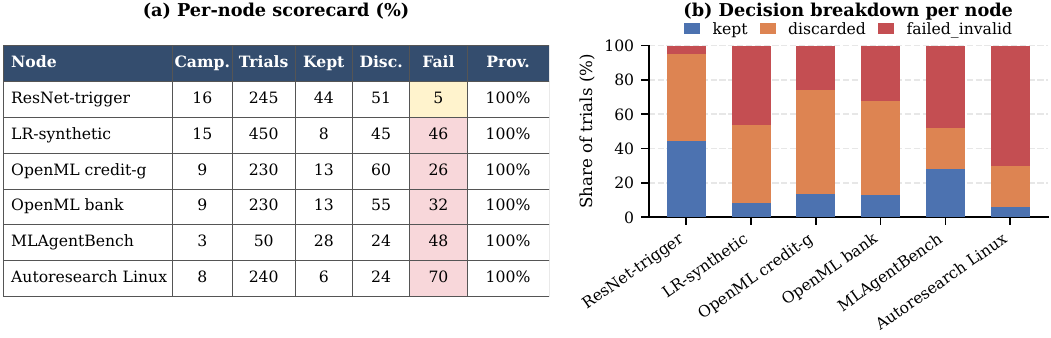}
\Description{Per-node governance scorecard and stacked decision breakdown across all six benchmark nodes, each achieving 100\% provenance completeness.}
\caption{Governance overview across all six benchmark nodes (1{,}445 trials). (a)~Per-node scorecard: number of campaigns (Camp.), trials, and the shares of kept, discarded and failed-invalid trials, with provenance completeness (Prov.); the Fail share is shaded amber below 20\% and red at or above 20\%. (b)~The same decision shares as stacked bars. Provenance completeness is 100\% on every node regardless of task type, worker class, or failure rate---including the 120-trial total-failure arm on \texttt{autoresearch\allowbreak\_linux}.}
\label{fig:governance-overview}
\end{figure*}

{\sloppy

Suppose an LLM agent that tunes a training script reports that validation AUC rose by 0.05. The score does not say how the gain was obtained. The agent may have edited a file it was not supposed to touch, a run that crashed may have been dropped from the record, or the same bad idea may have been tried many times before one variant worked. For scientific ML, where an experiment is only as useful as the evidence behind it, these questions matter as much as the score: an autonomous experiment needs \emph{auditability}, a traceable and reproducible record of every decision and its evidence.

Current practice does not provide this. ML-agent benchmarks such as MLAgentBench and MLE-bench ask whether the agent improved the task~\cite{huang2024mlagentbench,chan2024mlebench}; agent systems such as AIDE keep their search state inside the same loop that makes the decisions~\cite{jiang2025aide}; and experiment trackers record runs after they exist, without governing which runs are attempted or how failures are classified~\cite{mlflow,wandb}. What is missing is a way to measure the experimental process itself, and a record of that process that the agent does not control.

We address both with a governed harness. Following the view of an agent as a model plus a harness~\cite{trivedy2026anatomy}, which together form a compound AI system~\cite{zaharia2024compound}, we separate three roles. A manager, typically an LLM, proposes one bounded change per trial; a worker applies it and runs training; and a control plane owns everything else: the trial's lifecycle, the check that the patch touches only editable files, metric parsing, the keep or discard decision, and one append-only ledger record per trial, written whatever the outcome. None of these is delegated to the agent, so the manager can be replaced without changing what is recorded. From the ledger we compute six governance metrics alongside the task metric: acceptance rate, invalid-action rate, repeated-bad rate, provenance completeness, artifact-reference completeness, and a failure taxonomy (Section~\ref{governance-metrics}).

We evaluate this contract on six ML tasks over 1{,}445 trials: synthetic logistic regression, two public OpenML tabular tasks, an MLAgentBench vectorization task, a ResNet classifier for a scientific trigger task, and GPT pretraining. Four findings stand out (Figure~\ref{fig:governance-overview} gives the per-node overview). First, the same contract applied unchanged across tasks, worker types and failure regimes. Second, the ledger exposed failures that a task score hides: a no-memory arm in which all 120 trials failed, a benchmark adapter whose bounded proposals ran out, and a reset failure that node-state hashes revealed, after which the affected campaigns were rerun. Third, a matched counterfactual shows what is lost without the control plane: a loop that logs only successful trials lost 43 of 50 trials, and a campaign in which every trial fails leaves an empty ledger, indistinguishable from one that never ran (Section~\ref{stress-campaigns}). Complete provenance, which the design guarantees (Section~\ref{trial-lifecycle}), held in every trial. Fourth, memory ablations used as diagnostic probes show that memory's effect on repeated bad proposals depends on the task: summary memory gave the lowest repeated-bad rate in two of three ResNet replicates, but not on the synthetic task (Section~\ref{memory-ablation}).

Our contributions are:
\begin{itemize}
\tightlist
\item an evaluation protocol that reports governance metrics next to task metrics, with each metric assigned by the control plane rather than reported by the agent (Section~\ref{evaluation-protocol});
\item a control-plane design that separates proposal, execution, validation and keep/discard authority, so that the LLM manager is a replaceable component (Section~\ref{governed-control-plane});
\item an empirical test across six tasks and 1{,}445 trials, with the ledgers of all 1{,}445 trials released so that every number reported for them can be recomputed (Section~\ref{results} and Appendix~\ref{sec:supp-artifact}).
\end{itemize}

Section~\ref{background-and-related-work} reviews related work, Section~\ref{governed-control-plane} describes the control plane, Section~\ref{evaluation-protocol} gives the evaluation protocol, Section~\ref{results} reports the results, and Section~\ref{discussion-and-limitations} discusses limitations.\par}

\section{Background and Related Work}\label{background-and-related-work}

The key distinction of this paper is that we evaluate the \emph{harness} that governs the agent loop---lifecycle states, scope enforcement, provenance, and failure classification---not only the model that proposes edits.

\textbf{ML-agent benchmarks and code-space exploration.}
AIDE~\cite{jiang2025aide} is the closest academic system: both use LLMs to
propose atomic code changes evaluated by a hard-coded metric, but they differ
fundamentally in where lifecycle authority lives. AIDE is a coupled optimisation
loop --- search policy, evaluator, and solution summarisation share state inside
a single process --- and keeps its solution tree in memory. In the released
code (commit \texttt{ad73d22}, checked in November 2026)~\cite{aideml}, the main loop saves a snapshot of the whole journal after
every step, so a crash loses only the step in progress; but the snapshot is
overwritten rather than appended, there is no pending-trial guard, and a failed
node carries a buggy flag (set from an LLM review, an exception, or a missing
metric) and the exception type rather than a control-plane failure category.
AIDE generates whole solution scripts, so it has no editable-path whitelist or
scope enforcement. The present work inverts this
design: the control-plane contract is fixed and externally visible; the manager
and worker are replaceable without altering governance; every outcome---whether
kept, discarded, or crashed---produces an immutable typed audit record before the
next trial begins. The editable-path whitelist is a precondition guard, not a
post-hoc filter.

MLAgentBench~\cite{huang2024mlagentbench}, MLE-bench~\cite{chan2024mlebench},
FML-bench~\cite{zou2025fmlbench}, and MLGym~\cite{nathani2025mlgym} address a
different evaluation target: they ask \emph{did the agent improve performance on
this task?} This paper asks \emph{did the harness guarantee that every trial was
bounded, classified, and provenance-tracked, regardless of outcome?} The two
questions are complementary rather than competing. This paper wraps an
MLAgentBench-style task (NumPy vectorization speed) as the
\texttt{mlagentbench\_vec.} node, demonstrating that any task expressible as a
NodeSpec receives the same governance coverage, but the governance evaluation is
independent of task-metric outcome. SHARP~\cite{birk2026sharp} governs
reproduction of \emph{existing} experiments; this paper governs autonomous
\emph{generation} of new ones.

\textbf{Experiment tracking, AutoML, and NAS.}
Trackers such as MLflow and Weights \& Biases~\cite{mlflow,wandb} record outcomes after a run exists but do not govern the agent loop or classify invalid edit attempts before training. AutoML and neural architecture search systems~\cite{hutter2019automl,elsken2019nas} optimise within predefined search spaces; they are not designed to audit self-directed exploration, explain why a proposed change was kept or discarded, or expose lifecycle diversity. This paper makes the opposite tradeoff: a narrower set of controlled nodes with complete per-trial provenance.

\textbf{Harness engineering and governance.}
This framing is consistent with practitioner writing on harness engineering~\cite{trivedy2026anatomy,bockeler2026coders,hashimoto2026journey,openai2026codex,anthropic2025longrunninga}: Trivedy frames the harness as everything around the model, including tools, sandboxes, memory, middleware, and execution hooks. Böckeler argues that trust in agent output requires constraining the solution space with computational guides and feedback sensors. This paper is not another task benchmark; it is a governance layer into which any benchmark task can be wrapped via a NodeSpec contract. The node specification is a computational guide; the pending-trial guard and state machine are computational sensors; memory injection is an inferential guide; the repeated-bad detector is an inferential sensor. Agent memory and scaffolding frameworks~\cite{shinn2023reflexion,du2026memory} inform the design of context-resident memory injection and reflective self-improvement, but do not provide an evaluation protocol for governed autonomous experimentation.

\section{Governed Control Plane}\label{governed-control-plane}

The system is a governed harness for autonomous ML experimentation. The manager proposes an experimental change; the worker materialises a patch and run artifacts. The control plane validates scope, executes the lifecycle, parses metrics, assigns a kept, discarded, or failed-invalid decision, updates memory, and appends the authoritative ledger record. Managers and workers are replaceable components; the control plane contract is fixed.

\paragraph*{A worked trial.}
Before the full abstraction stack, we trace one governed trial end to end, drawn from the ResNet-trigger case study of Section~\ref{main-campaign-governance}.
(i)~\emph{Proposal:} the manager emits one bounded change --- summary, rationale, target parameter, intended effect --- and the control plane assigns it a proposal provenance ID.
(ii)~\emph{Scope validation:} the worker materialises a patch touching only \texttt{train.py}, the single whitelisted file; a patch touching frozen data or split logic would terminate here as \texttt{invalid\_edit\_scope}, before any training runs.
(iii)~\emph{Execution under guard:} the control plane writes a pending-guard file, then the worker runs training; a crash here would leave a stale guard that forces a \texttt{failed\_invalid} record at the next campaign start rather than a silent gap.
(iv)~\emph{Metric parse and decision:} the parser extracts \texttt{val\_auc} and the control plane --- not the manager --- compares it against the current best (this campaign improves 0.7747\,$\rightarrow$\,0.7827 over 20 trials); an improvement is \texttt{kept}, a valid non-improvement \texttt{discarded}.
(v)~\emph{Ledger append:} exactly one immutable JSONL record is appended --- proposal text, patch reference and hash, raw-log reference, parsed metric, decision, failure category if any, and five provenance IDs (proposal, patch, run, metric, and decision IDs, each derived by hashing the trial ID) --- the guard is deleted, and the next budget slot may open.
All 1{,}445 trials in this paper, including all failures, follow exactly this path; the only variation is which guard or comparison produces the terminal state.

To make auditability concrete, we operationalise it through the five desiderata in Table~\ref{tab:audit-desiderata}. These are not claimed to exhaust every possible failure mode of agentic experimentation, nor to be statistically independent axes. They define the minimum operational conditions needed for the claims in this paper to be inspectable. In particular, complete attempt accounting and provenance linkage are related but distinct: the former asks whether every attempted budget slot appears in the ledger, while the latter asks whether each recorded attempt carries enough evidence to inspect it.

\begin{table}[t]
\footnotesize
\caption{Five auditability desiderata for governed agentic experimentation.}
\label{tab:audit-desiderata}
\begin{tabular}{@{}L{0.34\linewidth}L{0.56\linewidth}@{}}
\toprule
Desideratum & Operational meaning \\
\midrule
D1. Bounded edit scope & The agent cannot change protected benchmark, data, or evaluation files. \\
D2. Complete attempt accounting & Every proposed attempt is recorded, including invalid proposals and crashes. \\
D3. Independent adjudication & Acceptance or rejection is decided by fixed control-plane rules, not by the agent. \\
D4. Provenance linkage & Each recorded attempt links to patch, configuration, run, metric, and decision evidence. \\
D5. Recomputability & Aggregate governance claims can be recomputed from the ledger without re-running training. \\
\bottomrule
\end{tabular}
\end{table}

\begin{figure}[t]
\centering
\resizebox{\columnwidth}{!}{%
\begin{tikzpicture}[
  font=\sffamily\small,
  >=Latex,
  repl/.style={draw, rounded corners=3pt, fill=gray!9, align=center,
               inner sep=5pt},
  fixd/.style={draw=blue!55, fill=blue!7, rounded corners=4pt,
               line width=1.1pt, align=center, inner sep=5pt},
  stor/.style={draw, rounded corners=3pt, fill=green!8, align=center,
               inner sep=5pt},
  evls/.style={draw, rounded corners=3pt, fill=orange!6, align=center,
               inner sep=5pt},
  state/.style={draw, rounded corners=6pt, align=center, inner sep=3pt,
               font=\sffamily\scriptsize, minimum width=1.45cm, fill=blue!6},
  arr/.style={->, line width=0.65pt},
  lbl/.style={font=\sffamily\scriptsize, align=center},
]

\node[repl, minimum width=3.2cm, minimum height=2.2cm] (mgr) at (2.0, 7.8) {
  \textbf{Manager}\\[1pt]
  \textit{\tiny replaceable}\\[3pt]
  \texttt{prompt\_manager}\\
  \texttt{LangGraph}\\
  \texttt{baseline}
};

\node[fixd, minimum width=3.2cm, minimum height=3.4cm] (cp) at (2.0, 4.2) {
  \textbf{Control Plane}\\[1pt]
  \textit{\tiny fixed contract}\\[3pt]
  scope validation\\
  lifecycle state machine\\
  keep\,/\,discard decision\\
  provenance ID assignment\\
  memory context update
};

\node[repl, minimum width=2.8cm, minimum height=1.8cm] (wkr) at (6.4, 5.6) {
  \textbf{Worker}\\[1pt]
  \textit{\tiny replaceable}\\[3pt]
  \texttt{LocalWorker}\\
  \texttt{ClawWorker}
};

\node[repl, minimum width=2.8cm, minimum height=2.4cm] (nod) at (6.4, 2.8) {
  \textbf{Node}\\[1pt]
  \textit{\tiny NodeSpec YAML}\\[3pt]
  \texttt{resnet}, \texttt{lr\_synth}\\
  \texttt{openml\_*}, \texttt{mlab.}\\
  \texttt{autoresearch}
};

\node[stor, minimum width=3.2cm, minimum height=2.0cm] (led) at (2.0, 0.4) {
  \textbf{Append-only Ledger}\\[1pt]
  \textit{\tiny JSONL --- never overwritten}\\[3pt]
  \texttt{trial\_id} $\cdot$ \texttt{patch\_ref}\\
  \texttt{decision} $\cdot$ \texttt{provenance}
};

\node[evls, minimum width=3.2cm, minimum height=1.5cm] (evl) at (2.0, -2.2) {
  \textbf{Evaluation \& Reporting}\\[3pt]
  acceptance $\cdot$ IR $\cdot$ RBR\\
  provenance $\cdot$ failure taxonomy
};

\draw[arr] (mgr.south) --
  node[lbl, right=2pt] {proposal}
  (cp.north);

\draw[arr] ($(cp.east)+(0,0.4)$) --
  node[lbl, above, sloped, pos=0.5] {worker packet}
  (wkr.west);

\draw[arr] (wkr.south) --
  node[lbl, right=2pt] {patch\,+\,log}
  (nod.north);

\draw[arr] (nod.west) --
  node[lbl, above, sloped, pos=0.5] {parsed metric}
  ($(cp.east)+(0,-0.4)$);

\draw[arr] (cp.south) --
  node[lbl, right=2pt] {one record per trial}
  (led.north);

\draw[arr] (led.south) --
  node[lbl, right=2pt] {export}
  (evl.north);

\draw[arr] (led.west)
  .. controls (-1.8, 0.4) and (-1.8, 7.8) ..
  node[lbl, left=3pt] {memory\\context}
  (mgr.west);

\node[draw=gray!55, dashed, rounded corners=4pt,
      fit=(mgr), inner xsep=5pt, inner ysep=4pt] {};
\node[draw=gray!55, dashed, rounded corners=4pt,
      fit=(wkr)(nod), inner xsep=5pt, inner ysep=4pt] {};

\node[draw=gray!55, dashed, rounded corners=4pt,
      minimum width=5.6cm, minimum height=3.0cm,
      label={[font=\sffamily\scriptsize, text=gray!65]above:trial lifecycle state machine}]
      (smbox) at (6.9, -0.85) {};

\node[state] (pending)   at (5.45,  0.15) {pending};
\node[state] (executing) at (5.45, -1.0) {executing};
\node[state] (kept)      at (8.5,   0.05) {kept};
\node[state] (disc)      at (8.5,  -1.0) {discarded};
\node[state, font=\sffamily\tiny, minimum width=1.65cm]
             (fail)      at (8.5,  -1.95) {\texttt{failed\_}\\\texttt{invalid}};

\draw[arr] (pending) --
  node[lbl, left=2pt] {pending\\guard}
  (executing);
\draw[arr] (executing.east) --
  node[lbl, above, sloped, pos=0.55, font=\sffamily\tiny] {improves}
  (kept.west);
\draw[arr] (executing.east) --
  node[lbl, above, pos=0.55, font=\sffamily\tiny] {valid,\,worse}
  (disc.west);
\draw[arr] (executing.east) --
  node[lbl, below, sloped, pos=0.55, font=\sffamily\tiny] {scope\,/\,crash}
  (fail.west);

\end{tikzpicture}%
}
\caption{System architecture and trial lifecycle. \emph{Replaceable components}
(manager, worker, node) are interchangeable without changing the governance
contract; the control plane contract is fixed across all six nodes and every
manager backend tested in this paper. The control plane owns scope validation,
lifecycle transitions, keep/discard decisions, provenance ID assignment, and
memory context updates. Inset (bottom right): the trial lifecycle state machine
(Section~\ref{trial-lifecycle}).}
\Description{Block diagram showing a manager sending a proposal to the control
plane, which dispatches a worker packet to the worker. The worker materialises
a patch on a node. The node returns a parsed metric to the control plane. The
control plane appends one immutable record per trial to the append-only ledger.
The ledger provides a memory context feedback arrow back to the manager, and
exports governance metrics to the evaluation and reporting layer. Manager,
worker, and node are grouped as replaceable components; the control plane has a
heavier blue border indicating a fixed contract. A dashed inset in the bottom
right shows the trial lifecycle state machine: pending transitions to executing
on the pending guard, then branches to kept (improves), discarded (valid but
worse), or failed invalid (scope violation or crash).}
\label{fig:architecture}
\end{figure}
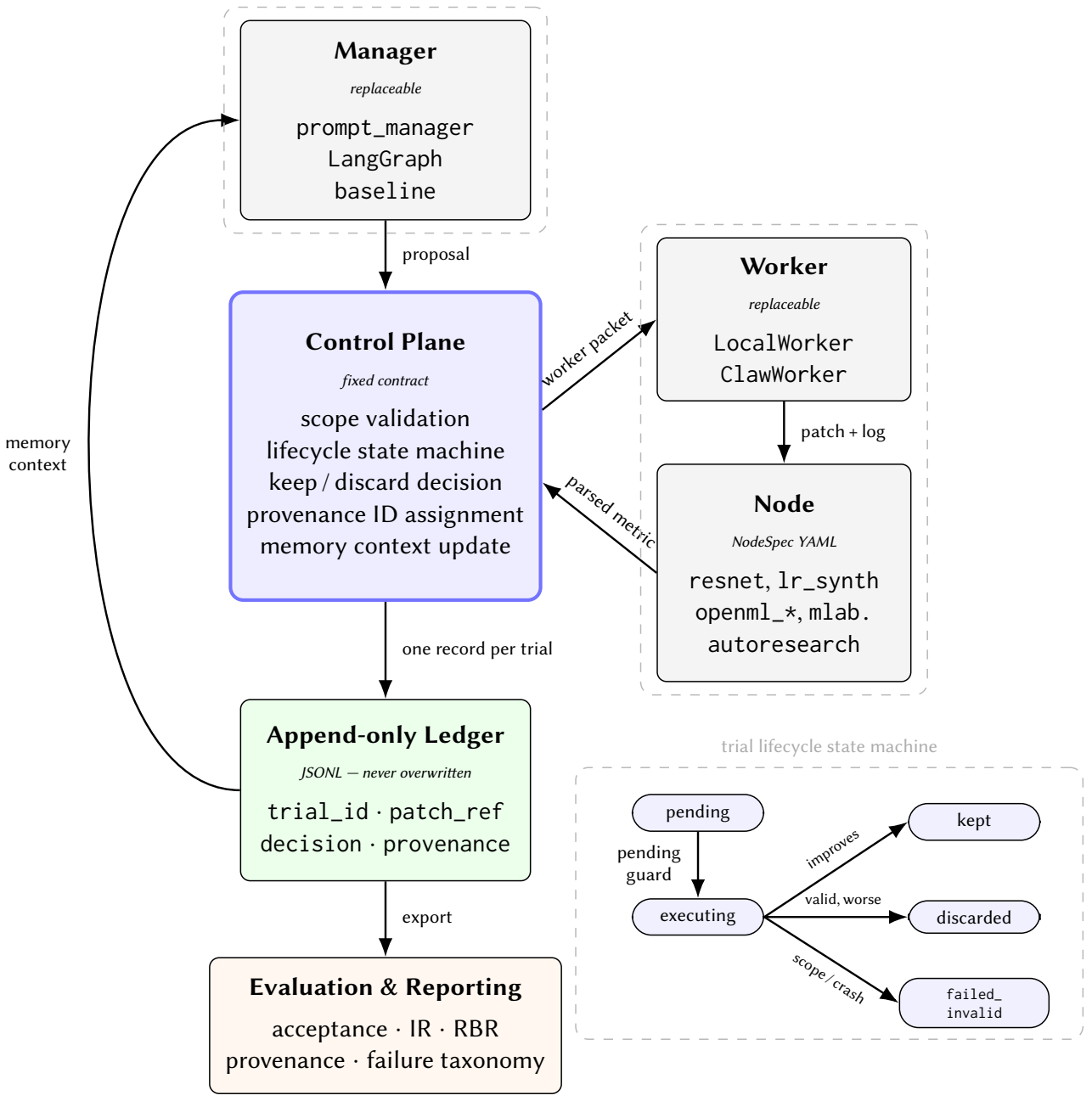

\subsection{Alignment with Agent Evaluation Criteria}\label{alignment-with-agent-evaluation-criteria}

Table~\ref{tab:kdd-alignment} maps the Evaluation and Trustworthiness of Agentic AI workshop concerns to the harness mechanisms and metrics reported in this paper.

\begin{table}[t]
\footnotesize
\caption{Workshop evaluation concerns mapped to harness mechanisms.}
\label{tab:kdd-alignment}
\begin{tabular}{@{}L{0.28\linewidth}L{0.38\linewidth}L{0.24\linewidth}@{}}
\toprule
Evaluation concern & Harness mechanism & Reported metric \\
\midrule
Repeated poor behaviour & Append-only memory + repeated-bad detector & Repeated-bad rate \\
No ground truth for intermediate choices & Decision authority separated from manager; provenance IDs per trial & Provenance completeness \\
Deployment and scope safety & NodeSpec whitelist, no-op guard, pending-trial guard & Invalid-action rate \\
Long-horizon failure accumulation & State machine; terminal \texttt{kept}/\texttt{disc.}/\texttt{failed\_inv.} & Failure taxonomy \\
Benchmark reproducibility & NodeSpec YAML, deterministic patch bridge, seeded replicates & Artifact manifest \\
\bottomrule
\end{tabular}
\end{table}

\subsection{Manager, Worker, and Replacement Principle}\label{managerworker-separation-and-replacement-principle}

The manager generates proposals (summary, rationale, target, intended effect). The worker executes bounded actions (patch + run artifacts). The control plane owns authority: validity, state transitions, budget, memory updates, provenance IDs, and final decision. This separation is the \emph{replacement principle}: any manager backend --- baseline heuristic, \texttt{prompt\allowbreak\_manager}, \texttt{LangGraph\-Manager}, or an AIDE-style tree-search --- can be swapped without changing the governance contract. Two worker implementations span the reproducibility spectrum: \texttt{Local\-Worker} applies ``Change PARAM from X to Y'' directives with no external dependencies; \texttt{Claw\-Worker} invokes an external coding-agent path for free-form proposals, routing structured proposals through a deterministic constant-patch bridge that requires only Ollama~\cite{ollama}.

\subsection{Benchmark Node Inventory}\label{node-inventory}

Table~\ref{tab:node-inventory} lists the six nodes reported in this paper. The control-plane contract is identical across all nodes: the only per-node configuration is the NodeSpec YAML (editable paths, frozen paths, run command, metric parser, and budget range).

\begin{table*}[t]
\footnotesize
\caption{The six benchmark nodes. All nodes share the same control-plane contract; only the NodeSpec YAML differs. Direction: max = maximise metric, min = minimise.}
\label{tab:node-inventory}
\begin{tabular}{@{}p{0.16\textwidth}p{0.14\textwidth}p{0.07\textwidth}p{0.10\textwidth}cp{0.40\textwidth}@{}}
\toprule
Node & Task type & Worker & Metric & Dir & Purpose in evaluation \\
\midrule
\texttt{resnet\_trigger} & Waveform classification & Claw & \texttt{val\_auc} & max & Primary case study; memory ablation substrate \\
\texttt{lr\_synthetic} & Logistic regression & Local & \texttt{val\_score} & max & Cross-node transfer check; dependency-free \\
\texttt{openml\_credit\_g} & Tabular classification & Local & \texttt{val\_auc} & max & Governance transfer to public tabular node \\
\texttt{openml\_bank\_mkt} & Tabular classification & Local & \texttt{val\_auc} & max & Governance transfer with visible task improvement \\
\texttt{mlagentbench\_vec.} & NumPy convolution & Local & \texttt{speed\_score} & max & Benchmark compatibility probe \\
\texttt{autoresearch\_linux} & LM pretraining & Claw & \texttt{val\_bpb} & min & Real-task stress case; governance under high failure rate \\
\bottomrule
\end{tabular}
\end{table*}

\subsection{Trial Lifecycle}\label{trial-lifecycle}

Each trial follows the fixed lifecycle shown in Figure~\ref{fig:architecture} (state machine inset). Every opened budget slot must produce a \mbox{\texttt{kept},} \mbox{\texttt{discarded},} or \texttt{failed\_invalid} audit object.

The control plane writes a pending guard before worker execution, making mid-trial crashes auditable. A scope validator checks the patch against the NodeSpec whitelist before training begins: patches touching frozen files are rejected, recorded, and prevented from corrupting node state. The append-only JSONL ledger stores, per trial: proposal summary and rationale, patch and log references, parsed metrics, validity status, decision, failure category, provenance IDs, and reproducibility hashes.

Table~\ref{tab:state-machine} formalises the lifecycle.

\begin{table}[t]
\footnotesize
\caption{Trial lifecycle state machine. The pending guard is a file written before \textsc{executing} and deleted on any terminal transition; its presence at campaign start raises \texttt{PendingTrialError}, preventing concurrent writes to the append-only ledger.}
\label{tab:state-machine}
\begin{tabular}{@{}L{0.18\linewidth}L{0.20\linewidth}L{0.52\linewidth}@{}}
\toprule
From state & To state & Condition \\
\midrule
\textsc{pending} & \textsc{executing} & Pending guard written; worker starts \\
\textsc{executing} & \textsc{kept} & Metric parses; beats current best; patch diff retained \\
\textsc{executing} & \textsc{discarded} & Metric parses; does not beat current best \\
\textsc{executing} & \textsc{failed\_invalid} & Scope violation, runtime error, missing metric, no-op patch, or precondition failure \\
\textsc{executing} & \textsc{failed\_invalid} & Mid-trial crash detected via stale pending guard at next campaign start \\
\midrule
\multicolumn{3}{@{}p{\linewidth}@{}}{\textit{Terminal states:} \textsc{kept}, \textsc{discarded}, \textsc{failed\_invalid}. All three produce a complete ledger record. Recovery tool: \passthrough{\lstinline!scripts/recover\_pending.py!}.} \\
\bottomrule
\end{tabular}
\end{table}

\paragraph*{Lifecycle invariant.}
A trial begins in \textsc{pending}, then enters \textsc{executing}. It ends in
exactly one terminal state: \textsc{kept}, \textsc{discarded}, or
\textsc{failed\_invalid}. The control plane maintains the following invariant:

\begin{quote}\small\sloppy
\emph{For every budget slot that enters \textsc{executing}, the control plane
appends exactly one terminal record---\textnormal{\texttt{kept}}, \textnormal{\texttt{discarded}}, or
\textnormal{\texttt{failed\_invalid}}---to the append-only ledger before the next slot may
open. No slot is silently skipped; no slot produces more than one record;
no record is overwritten or deleted.}
\end{quote}

Enforcement has two paths. \textit{Normal path:} the pending guard is written
before the worker starts and deleted on the terminal transition; any deviation
raises an error before the ledger append. \textit{Crash-recovery path:} if the
process crashes mid-trial, the guard file persists; at the next campaign start
\texttt{PendingTrialError} is raised, the stale guard is inspected via
\texttt{scripts/recover\_pending.py}, and a \texttt{failed\_invalid} record is
appended before any new trial begins. Neither path allows a budget slot to close
without a ledger record.

The invariant is structural rather than empirical: it holds by construction of
the control-plane code, not because nothing crashed during the reported
experiments. Complete provenance is accordingly observed in all 1{,}445
trials, as the invariant requires.

\subsection{Failure Taxonomy and Metric Validation}\label{failure-taxonomy}

Failure labels are assigned by the control plane at the terminal transition that writes the ledger record, not post hoc by the manager. The reported evidence uses five observed \texttt{failed\allowbreak\_invalid} categories: \texttt{invalid\allowbreak\_edit\allowbreak\_scope}, \texttt{no\allowbreak\_op\allowbreak\_patch}, \texttt{proposal\allowbreak\_precondition\allowbreak\_failed}, \texttt{runtime\allowbreak\_error}, and node-level \texttt{invalid\allowbreak\_config}. The schema also supports \texttt{metric\allowbreak\_missing} and \texttt{edit\allowbreak\_failed}; those labels do not appear in the primary 1{,}445-trial evidence set. Valid non-improvements receive \texttt{discarded}, not \texttt{failed\allowbreak\_invalid}; the label \texttt{degraded\allowbreak\_metric} on their records marks a discarded trial, not a failure. The taxonomy is exhaustive for the current NodeSpec contract because every terminal transition is produced by one of those guards or by a parsed metric comparison. New nodes may add subcategories only if they map to the same three terminal outcomes before the ledger append. Structural validation uses forced stress campaigns, paired deterministic reruns, kept-artifact re-execution, and rule-based relabeling from observable ledger fields. Readers who want the taxonomy grounded before the aggregate results should consult Appendix~\ref{sec:supp-failure-taxonomy}, which gives one concrete ledger example per category with its detection point and guard type.

\subsection{Memory Modes and Decision Authority}\label{memory-modes}

The memory interface has three modes: no prior-trial context, append-only summary, and rationale-augmented summary. The rationale arm adds per-trial failure rationales and can truncate them with \passthrough{\lstinline!rationale\_max\_tokens!}. The ablation is a diagnostic probe: it tests whether the manager changes behavior when exposed to prior governed outcomes, and whether repeated-bad rate can detect that behavioral change. It is not treated as evidence that any memory format is generally superior.

The keep/discard decision lives in the control plane, not the manager. A valid candidate is kept only if the configured metric improves over the current best; a valid non-improving candidate is \passthrough{\lstinline!discarded!}; an invalid candidate is \passthrough{\lstinline!failed\_invalid!} with a specific failure category. This deterministic, externally-owned decision rule follows practitioner guidance for long-running agents: harnesses should provide review loops and external checks rather than letting the model own success criteria~\cite{anthropic2026longrunningb}.

\section{Evaluation Protocol}\label{evaluation-protocol}

The evaluation asks whether an autonomous ML experimentation loop is bounded, auditable, failure-aware, reproducible, and behaviorally affected by governance memory. Governance metrics are primary; task-metric AUC is secondary evidence that the loop runs end-to-end.

\subsection{Benchmark Nodes}\label{benchmark-node}

The primary scientific benchmark is \passthrough{\lstinline!resnet\_trigger!}, a near-threshold detector waveform classification task with frozen H5 corpora, split logic, preparation code, and node metadata; workers may edit only \passthrough{\lstinline!nodes/ResNet\_trigger/train.py!}. The synthetic transfer benchmark, \passthrough{\lstinline!lr\_synthetic!}, is a pure-NumPy logistic-regression classifier run via \passthrough{\lstinline!LocalWorker!} for dependency-free protocol validation; its data-generation constants are frozen.

\begin{sloppypar}
We also instantiate two public OpenML~\cite{vanschoren2013openml} tabular nodes: \texttt{openml\allowbreak\_credit\allowbreak\_g} (dataset 31, the Statlog German Credit data~\cite{hofmann1994statlog}) and \texttt{openml\allowbreak\_bank\allowbreak\_marketing} (dataset 1461, the Bank Marketing data~\cite{moro2014bank}). They expose only bounded \passthrough{\lstinline!config.yaml!} hyperparameters and model choices; dataset identity, target column, split logic, preprocessing, parser, and NodeSpec metadata are frozen. These are governance-transfer substrates, not competitive AutoML benchmarks.
\end{sloppypar}

The \passthrough{\lstinline!mlagentbench\_vectorization!} node wraps a NumPy-convolution speed task as a bounded external-benchmark adapter. Benchmark code, correctness checking, and parser logic are frozen; only \passthrough{\lstinline!config.yaml!} edits over implementation choice and workload size are exposed. This is a compatibility probe, not a full MLAgentBench benchmark claim.

The sixth node, \passthrough{\lstinline!autoresearch\_linux!}, is a GPT pretraining task on NVIDIA L40S hardware, adapted from Karpathy's autoresearch setup~\cite{karpathy2026autoresearch} through its macOS/CPU fork~\cite{andreenko2026autoresearchmacos}: a small GPT, with training code that simplifies nanochat~\cite{karpathy2025nanochat}, trained on the \texttt{climbmix-\allowbreak 400b-\allowbreak shuffle} data shards that the setup downloads. Its metric is \texttt{val\_bpb} (bits per byte, minimise), the paper's only minimise-direction task. Workers may edit only named hyperparameters in \passthrough{\lstinline!train.py!}, including the model's depth, aspect ratio, head dimension and attention window pattern as well as batch sizes and learning rates; data preparation, data, tokenizer and parser are frozen. Each trial takes approximately 6 minutes, so only \texttt{none} and \texttt{summary} arms are run: 4 independent seeds $\times$ 30 trials per arm = 240 trials.

\subsection{Campaign Protocol}\label{fixed-budget-campaign-protocol}

Controlled ablations are designed to start each campaign from a reset node state. The canonical scientific-node campaign uses \texttt{prompt\allowbreak\_manager} with structured state-aware hyperparameter proposals and \texttt{claw\allowbreak\_style\allowbreak\_worker} with the deterministic patch path. The memory ablation runs equal budgets across three memory modes (\texttt{none}, summary, and rationale-augmented summary) under the same node, budget, parser, and reset discipline. The OpenML campaigns use \texttt{LangGraph\-Manager}, summary memory, and 20- and 30-trial budgets on each public node. The MLAgentBench adapter uses \texttt{Local\-Worker} and a 30-trial budget. Stress campaigns exercise forced failure paths at 100\% rate. For every trial, the control plane records proposal, patch reference, raw log reference, parsed metrics, validity status, failure category, decision rationale, provenance IDs, and reproducibility hashes.

\textbf{Implementation and hardware environment.}
All cross-node campaigns use \texttt{deepseek/deepseek-v4-flash}~\cite{deepseek2026v4} as the proposal manager via LangGraph~\cite{langgraph}, holding the model fixed to reduce the capability confound across nodes. Model-agnostic findings include provenance completeness, lifecycle correctness, and scope enforcement; proposal-quality findings may be manager-specific. ClawWorker campaigns ran on a server-class NVIDIA L40S host (CUDA 12.8, Ubuntu 22.04, PyTorch 2.x); LocalWorker campaigns ran on a CPU-only host under Python 3.11. The corrected ResNet L40S replication resets all editable node files to a verified frozen baseline before trial~1; older \texttt{autoresearch\allowbreak\_linux} reset semantics are disclosed separately in Sections~\ref{autoresearch-linux} and~\ref{discussion-and-limitations}. Patch and run-log references are recorded with provenance IDs in the ledger; the ledgers needed to recompute the governance metrics are released with the code (Appendix~\ref{sec:supp-artifact}), while raw execution-host artifacts are not required for the paper's primary claims. Full environment details and campaign IDs are given in Appendices~\ref{sec:supp-reproducibility} and~\ref{sec:supp-tables}.

\subsection{Memory Ablation Design}\label{memory-ablation-design}

As a diagnostic probe, we test the behavioral hypothesis
\[ \text{RBR}(\text{none}) > \text{RBR}(\text{summary}) > \text{RBR}(\text{rationale}) \]
where RBR is \emph{repeated-bad rate}: the fraction of proposals repeating the same edit target and mechanism as a prior rejected trial without a new corrective rationale. Rationale-linked memory should reduce repeated-bad proposals more than raw summaries because it provides targeted failure signals rather than history noise~\cite{shinn2023reflexion,wu2025metapolicy,openai2026codex,bockeler2026coders}.

Two manager configurations are tested on ResNet-trigger: a deterministic prompt manager (design-constrained negative --- round-robin makes memory structurally irrelevant) and a LangGraph manager calling qwen2.5-coder:7b~\cite{hui2024qwen25coder} at temperature 0.7 with an explicit AVOIDANCE RULE. All LangGraph proposals are parsed into structured parameter-change fields and applied via the deterministic patch bridge, eliminating coding-agent dependencies. Three ResNet-trigger replicates are run (two at budget 10, one at budget 20); a four-arm verbosity ablation with 50-token rationales tests whether rationale length drives the result. The same three-arm LangGraph ablation is repeated on \passthrough{\lstinline!lr\_synthetic!} as a cross-node transfer check.

\textbf{Large-scale replication (NVIDIA L40S).} To obtain variance estimates and test the hypothesised ordering, a third ablation configuration runs on an institutional GPU cluster: 3 memory modes $\times$ 5 independent seeds $\times$ 15 trials $= 225$ total trials. The manager is \texttt{deepseek-v4-flash}; the worker is \texttt{qwen2.5\allowbreak-coder\allowbreak:7b} at temperature~0.2. Each seed resets \texttt{train.py} before trial~1. Per-trial SHA-256 context hashes verify that memory accumulates across the 225-trial ledger.

\subsection{Governance Metrics}\label{governance-metrics}

\begin{table}[t]
\small
\caption{Governance metric suite. Given $N$ trials, these metrics characterise process reliability independent of task-metric outcome.}
\label{tab:governance-metrics}
\begin{tabular}{@{}p{0.28\linewidth}p{0.38\linewidth}p{0.24\linewidth}@{}}
\toprule
Metric & Formal definition & What it detects \\
\midrule
Acceptance rate (AR) & $|\text{kept}| / N$ & Valid improving edits \\
Invalid rate (IR) & $|\text{failed\_invalid}| / N$ & Unevaluable trials \\
Repeated-bad rate (RBR) & Repeated-bad proposals $/ N$ & Redundant failure modes \\
Provenance completeness & Trials with all five IDs $/ N$ & Records lacking the ID chain \\
Artifact-reference coverage & Patch/log refs and hashes for materialized runs; classified records for pre-run failures & Ledger-level evidence traceability \\
Failure taxonomy & Counts per \passthrough{\lstinline!FailureCategory!} & Classified, not hidden, failures \\
\bottomrule
\end{tabular}
\end{table}

Failure labels are control-plane-assigned. The categories observed in the reported evidence include: \texttt{runtime\allowbreak\_error}, \texttt{invalid\allowbreak\_edit\allowbreak\_scope}, \texttt{no\allowbreak\_op\allowbreak\_patch}, \texttt{proposal\allowbreak\_precondition\allowbreak\_failed}, and node-level \texttt{invalid\allowbreak\_config}. The schema also supports \texttt{metric\allowbreak\_missing} and \texttt{edit\allowbreak\_failed}, although those labels are not part of the primary 1{,}445-trial evidence set. Valid-but-non-improving trials receive \texttt{degraded\allowbreak\_metric} and are classified \texttt{discarded}, not \texttt{failed\allowbreak\_invalid}.
A campaign with high invalid rate but 100\% provenance completeness is more interpretable than one with unexplained score changes and no audit trail.

\subsection{Metric Validation Probes}\label{metric-validation-probes}

We run three lightweight validation probes for the governance metrics. First, two reruns of the MLAgentBench vectorization adapter test whether the same NodeSpec contract continues to produce complete lifecycle records under repeated runs. Second, a re-execution script rebuilds kept-trial state from cumulative patch references and reruns selected kept trials where the underlying execution artifacts are available. Third, we estimate structural relabeling agreement by computing Cohen's $\kappa$ between control-plane-assigned labels and an independent rule-based relabeler applied to structurally observable fields only (scope-validation flags, patch presence, training-initiation signals, execution status). On a stratified sample of 50 \texttt{failed\_invalid} trials spanning the observed validation categories, the structural relabeler achieves $\kappa = 0.830$ (95\%~CI~$[0.703,\,0.958]$). This is a structural consistency check, not a human inter-rater study. The remaining discordant records are edge cases where the ledger-level structural evidence does not distinguish the precise worker-side failure without consulting the referenced artifact file. Full human inter-rater validation remains future work.

\section{Results}\label{results}

\begin{sloppypar}
Results are presented in governance-first order. All numbers for the 1{,}445 primary trials derive from the primary campaign ledgers in \passthrough{\lstinline!experiments/ledgers/!}, which are released with the code (Appendix~\ref{sec:supp-artifact}); development and smoke ledgers are excluded from the paper totals. All governance metrics (acceptance rate, invalid rate, repeated-bad rate, provenance completeness) are recomputable from these ledgers without re-running any training; no GPU is required to verify the primary claims. The primary paper-relevant campaign set comprises 1{,}445 governed trials, indexed in Appendix~\ref{sec:supp-campaign-ids}. Across the primary set, the harness runs one real-worker scientific case study (\texttt{resnet\_trigger}), one dependency-free synthetic transfer node (\texttt{lr\_synthetic}), two public OpenML tabular nodes, one MLAgentBench adapter, and one real LM training node (\texttt{autoresearch\allowbreak\_linux}); supplemental validation adds two one-trial stress campaigns. Together these campaigns exercise kept, discarded, failed-invalid, invalid-scope, no-op, invalid-configuration, and repeated-bad outcomes under a single ledger schema.
\end{sloppypar}

The section is staged as follows: the single-node case study (\S\ref{main-campaign-governance}); the six-node governance view (\S\ref{cross-node-summary}); stress campaigns and the governed-vs-ungoverned counterfactual (\S\ref{stress-campaigns}); the memory-ablation diagnostic and its large-scale replication (\S\S\ref{memory-ablation}--\ref{large-scale-replication}); and the real LM-training stress case (\S\ref{autoresearch-linux}). Throughout, AR denotes acceptance rate, IR invalid rate, and RBR repeated-bad rate (Table~\ref{tab:governance-metrics}).

\subsection{Real-Worker Scientific Node Case Study}\label{main-campaign-governance}

Table~\ref{tab:main-campaign} summarises the 20-trial real-worker campaign on ResNet-trigger. Four valid improving edits are kept, five valid non-improvements are discarded with patch/log references and metric evidence retained in the ledger, and eleven impossible/no-op structured proposals are classified as \texttt{proposal\allowbreak\_precondition\allowbreak\_failed} before they can corrupt node state. Provenance is complete for all 20 records (100\%).

\begin{table}[t]
\footnotesize
\caption{Real-worker scientific-node case study (\texttt{kdd\_resnet\_scientific\_20}, ResNet-trigger, \texttt{prompt\_manager}, \texttt{claw\_style\_worker}, budget=20).}
\label{tab:main-campaign}
\begin{tabular}{@{}p{0.40\linewidth}p{0.55\linewidth}@{}}
\toprule
Field & Value \\
\midrule
Kept / disc. / failed-invalid & 4 / 5 / 11 \\
Non-kept categories & 5 $\times$ \texttt{degraded\allowbreak\_metric}; 11 $\times$ \texttt{proposal\allowbreak\_precondition\allowbreak\_failed} \\
Acceptance / invalid rate & 0.20 / 0.55 \\
Provenance completeness & 1.00 (20/20) \\
Decision record completeness & 1.00 (20/20) \\
Failure evidence completeness & 1.00 (11/11 precondition failures classified) \\
Patch refs/hashes for materialized valid runs & 1.00 (9/9 valid runs) \\
Initial / best \texttt{val\_auc} & 0.7747 / 0.7827 (+0.008022) \\
Baseline seed noise (5 seeds) & mean 0.785, bootstrap 95\% CI of mean [0.775, 0.798] \\
\bottomrule
\end{tabular}
\end{table}

The net AUC gain (+0.008022) remains within the baseline seed interval and should be treated as secondary evidence that the loop runs correctly, not as an optimisation claim.

\subsection{Cross-Node Governance Transfer}\label{cross-node-summary}

\begin{figure}[t]
\centering
\includegraphics[width=\linewidth]{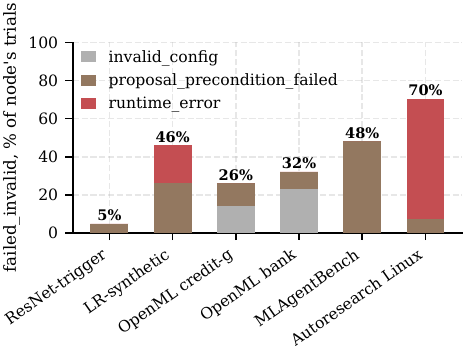}
\Description{Stacked bar chart showing the proportion of each FailureCategory within failed-invalid outcomes per benchmark node.}
\caption{Failure taxonomy by node: \texttt{failed\_invalid} trials as a share of each node's trials, split by \texttt{FailureCategory}. Composition is node-specific: \texttt{autoresearch\allowbreak\_linux} concentrates in \texttt{runtime\_error}; tightly-constrained nodes accumulate \texttt{proposal\_precondition\_failed}.}
\label{fig:ir-taxonomy}
\end{figure}

Table~\ref{tab:cross-node-summary} and Figure~\ref{fig:governance-overview} consolidate all six nodes into a single governance view; Figure~\ref{fig:ir-taxonomy} breaks the failed-invalid trials down by failure category. The central finding is that provenance completeness is 100\% on every node without exception --- including the 120-trial none-arm total failure and the s2 anomaly on \texttt{autoresearch\allowbreak\_linux}, and across all three lifecycle outcomes (kept, discarded, failed-invalid).

\begin{table*}[t]
\footnotesize
\caption{Cross-node governance summary. All nodes achieve 100\% provenance completeness regardless of task type, worker class, or failure rate. AR = acceptance rate, IR = invalid rate. ResNet-trigger appears as two rows because the campaign configurations differ fundamentally (prompt\_manager vs.\ DeepSeek LangGraph; different memory modes). The autoresearch arms are listed separately because their AR/IR differ categorically. The \texttt{lr\_synthetic} row pools its three memory modes.}
\label{tab:cross-node-summary}
\begin{tabular}{@{}p{0.31\textwidth}rrrp{0.42\textwidth}@{}}
\toprule
Node & Trials & AR\% & IR\% & Lifecycle outcomes \\
\midrule
\texttt{resnet\_trigger} (case study) &  20 & 20.0 & 55.0 & kept, disc., failed\_inv. \\
\texttt{resnet\_trigger} (L40S, all arms) & 225 & 46.7 &  0.4 & kept, disc., failed\_inv. (5 clean seeds verified) \\
\texttt{lr\_synthetic}           & 450 &  8.4 & 46.2 & kept, disc., failed\_inv. \\
\texttt{openml\allowbreak\_credit\allowbreak\_g}       & 230 & 13.5 & 26.1 & kept, disc., failed\_inv. \\
\texttt{openml\allowbreak\_bank\allowbreak\_marketing} & 230 & 13.0 & 32.2 & kept, disc., failed\_inv. \\
\texttt{mlagentbench\_vec.}      &  50 & 28.0 & 48.0 & kept, disc., failed\_inv. \\
\texttt{autoresearch} (none arm) & 120 &  0.0 & 100.0 & failed\_inv.\ only (total failure) \\
\texttt{autoresearch} (summary)  & 120 & 11.7 & 40.8 & kept, disc., failed\_inv. \\
\midrule
\multicolumn{5}{@{}p{0.80\textwidth}@{}}{\textbf{Provenance completeness: 100\% on every row above}} \\
\bottomrule
\end{tabular}
\end{table*}

\begin{sloppypar}
\textbf{Public tabular nodes.} \texttt{openml\allowbreak\_credit\allowbreak\_g} and \texttt{openml\allowbreak\_bank\allowbreak\_marketing} test whether governed keep/discard/fail decisions transfer to public tabular substrates (LangGraph manager, summary memory, b20 and b30 budgets). Both nodes expose only bounded \texttt{config.yaml} edits; dataset, split, preprocessing, and parser code are frozen. Provenance completeness is 1.00 across all 460 trials combined. \texttt{openml\allowbreak\_bank\allowbreak\_marketing} shows visible task-metric improvement (mean best AUC 0.9295 at b20, 0.9341 at b30; 95\% CI\,[0.9337,\,0.9345] at b30, $\sigma{=}0.0005$), while \texttt{openml\allowbreak\_credit\allowbreak\_g} is a neutral transfer case (mean best AUC 0.7679 at b30). Invalid rates increase with budget as bounded candidates are exhausted or violate node-level configuration constraints; all such trials are classified as \texttt{proposal\allowbreak\_precondition\allowbreak\_failed} or \texttt{invalid\allowbreak\_config} rather than silently dropped.
\end{sloppypar}

\textbf{MLAgent\-Bench adapter.} The 30-trial main campaign records 2~kept, 4~discarded, and 24~failed-invalid trials with 1.00 provenance completeness. The high failed-invalid count reflects exhausted bounded candidates (\texttt{proposal\allowbreak\_precondition\allowbreak\_failed}), not benchmark failure; the control plane classifies them explicitly rather than inventing out-of-scope edits. Two additional 10-trial reruns add kept and discarded outcomes under the same NodeSpec contract, bringing the adapter evidence set to 50 trials with complete provenance.

\subsection{Stress Campaigns and Ungoverned Counterfactual}\label{stress-campaigns}

Table~\ref{tab:stress} documents forced failure paths and the synthetic-node governance validation.

\begin{table*}[t]
\scriptsize
\caption{Stress campaigns and synthetic-node validation. Each row is a silent failure in a naive unguarded loop that the harness makes visible.}
\label{tab:stress}
\begin{tabular}{@{}p{0.20\linewidth}p{0.18\linewidth}p{0.25\linewidth}p{0.27\linewidth}@{}}
\toprule
Scenario & Campaign & Governed outcome & Evidence \\
\midrule
Out-of-scope edit (frozen file touched) & \texttt{kdd\_stress\_scope} & Patch rejected before training; classified \texttt{invalid\_edit\_scope}; git state unchanged & 1/1 trials; patch ref retained \\
No-op patch (byte-identical edit) & \texttt{kdd\_stress\_noop} & Training skipped; classified \texttt{no\_op\_patch}; empty patch hash recorded & 1/1 trials; ledger record present \\
Synthetic baseline: \texttt{lr\_synthetic} (logistic regression, LocalWorker) & \texttt{lr\_synth\_baseline} & All three lifecycle outcomes in 5 trials; provenance complete; NodeSpec YAML portable & 2 kept, 1 discarded, 2 failed-invalid; best val\_score = 0.923 \\
Synthetic LangGraph transfer & \texttt{lr\_synth\_lg\_*} & Same manager/memory ablation runs on a different node and worker type; all data-generation constants remain frozen & 30/30 valid trials; RBR none/summary/rationale = 0.70/0.70/0.80 \\
\bottomrule
\end{tabular}
\end{table*}

The scope-violation and no-op cases are backed by real ledger records. The \passthrough{\lstinline!lr\_synthetic!} campaigns confirm that the governance protocol operates on a different ML task and worker type without code changes.

\textbf{Ungoverned counterfactual.}
Every ``the harness makes X visible'' claim is only as strong as evidence that X would be invisible without governance. Across the 1{,}445-trial primary campaign set, governance classification acts on the majority of outcomes: 242 trials (16.7\%) produce \passthrough{\lstinline!runtime\_error!} records that the pending-trial guard and append-only ledger make auditable; 220 trials (15.2\%) are caught by precondition checks before reaching a valid run; 85 trials (5.9\%) expose node-level \passthrough{\lstinline!invalid\_config!} failures; 662 valid-but-degrading trials (45.8\%) are discarded by the control-plane decision rule. Without the pending-trial guard and append-only ledger, each \passthrough{\lstinline!runtime\_error!} would be a silent filesystem event with no trial identifier, failure taxonomy, or provenance chain.

\begin{sloppypar}
Table~\ref{tab:counterfactual} reports matched governed/un\-governed pairs on two nodes chosen to span failure rates; the visual companion is Fig.~\ref{fig:supp-counterfactual}. The ungoverned arm is an independent matched campaign, not a reconstruction from the governed ledger. It disables the pending-trial guard, failure-record appends, and blocking scope enforcement (Table~\ref{tab:supp-validation-campaigns}). A separate observation log (\texttt{*\_ungoverned\allowbreak\_obs.jsonl}) records every executed trial and is the denominator for ledger completeness. Because failure records are not appended, the trials missing from the ungoverned ledger are, by construction, exactly that arm's failed trials: the comparison shows what such a loop loses, not an independent estimate of loss.
\end{sloppypar}

\begin{table*}[t]
\footnotesize
\caption{Governed vs.\ ungoverned counterfactual comparison across two nodes.
Ledger completeness $=$ (trials with a ledger record)\,/\,(trials that ran).
Governed completeness is 1.00 by structural invariant
(Section~\ref{trial-lifecycle}); ungoverned completeness falls below 1.00
because \texttt{failed\_invalid} trials are silently absent from the ledger;
by construction it equals one minus the ungoverned arm's failure rate.
IR\,=\,invalid rate of the governed arm.
95\% bootstrap CIs: 10{,}000 resamples, seed~42.}
\label{tab:counterfactual}
\begin{tabular}{@{}p{0.19\textwidth}rp{0.14\textwidth}p{0.16\textwidth}rp{0.28\textwidth}@{}}
\toprule
Node (profile) & $N$ & Governed & Ungoverned & Drops & Interpretation \\
\midrule
\texttt{autoresearch\allowbreak\_linux}
  \newline IR\,=\,100\% &
  30 &
  1.00 $[1.00,\,1.00]$ &
  0.00 $[0.00,\,0.00]$ &
  30\,(100\%) &
  Total failure: ungoverned ledger is empty; campaign indistinguishable from one that never ran \\[2pt]
\texttt{openml\allowbreak\_bank\allowbreak\_mkt}
  \newline IR\,=\,55\% &
  20 &
  1.00 $[1.00,\,1.00]$ &
  0.35 $[0.15,\,0.55]$ &
  13\,(65\%) &
  Mixed outcomes: ungoverned ledger retains only 7 valid trials; 13 classified failures leave no trace \\
\midrule
\multicolumn{6}{@{}p{\linewidth}@{}}{%
  \textit{Combined: 43\,/\,50 ungoverned trials absent, i.e.\ all failed trials of
  these two campaigns, which were chosen to span failure rates;
  0\,/\,50 governed trials absent (100\% completeness by structural invariant).}} \\
\bottomrule
\end{tabular}
\end{table*}

The two nodes bracket the failure spectrum. The \texttt{autoresearch\allowbreak\_linux} row is an extremal, definitional case: when every trial fails and failure appends are disabled, completeness is necessarily zero. The informative mixed-outcome comparison is \texttt{openml\allowbreak\_bank\allowbreak\_mkt}, where the ungoverned ledger retains only the 7 valid trials and silently omits all 13 failures. Combined, 43 of 50 ungoverned trials leave no ledger record versus 0 of 50 governed; because the two nodes were chosen to span failure rates, this combined fraction describes these campaigns, not a typical loss rate. The 95\% bootstrap CI for ungoverned completeness lies strictly below 1.00 on both nodes.

\subsection{Memory Ablation as Diagnostic Probe}\label{memory-ablation}

Memory context is evaluated as a diagnostic probe for governance behaviour, not as a proposed optimisation method. The behavioral hypothesis is:
\[ \text{RBR}(\text{none}) > \text{RBR}(\text{summary}) > \text{RBR}(\text{rationale}) \]
where RBR is \emph{repeated-bad rate}: the fraction of proposals repeating the same edit target and mechanism as a prior rejected trial without a new corrective rationale.

Full arm-level results appear in Table~\ref{tab:supp-memory-ablation}. These ResNet-trigger campaigns are not among the 1{,}445 primary trials; their ledgers are in the archive of commit \texttt{62e93a8} rather than in the tagged release (Appendix~\ref{sec:supp-artifact}). We use this ablation only as a diagnostic example for repeated-bad rate, not as a proposed memory method.

\textbf{Primary diagnostic finding.} On ResNet-trigger, the summary arm has the lowest repeated-bad rate of the three arms in 2 of 3 replicates (rep1 and rep3, where the no-memory and rationale arms tie); in rep2 the rationale arm is lowest, so the full hypothesised ordering holds only in rep2 (Table~\ref{tab:supp-memory-ablation}). The node-state hashes show, however, that the rep2 arms were not reset between campaigns: each started from the final \texttt{train.py} of the arm run before it, so the rep2 rationale arm began from the summary arm's best state (\texttt{val\_auc} 0.7903 after its first trial). Rep2 is therefore not an independent replicate, and the one replicate in which the full ordering holds does not support it; rep1 and rep3 began every arm from the same node state. The verbosity arm C, run just before rep2, likewise began from the baseline plus one \texttt{BATCH\_SIZE} edit, which has no effect in this mode. These are descriptive counts over three short replicates, not a significance test. These ResNet-trigger runs used the node's CPU fast-search mode, in which the training-set size, trace length, batch size and number of epochs are taken from the environment; edits to \texttt{N\_SIGNAL}, \texttt{N\_NOISE}, \texttt{TRACE\_LEN}, \texttt{BATCH\_SIZE} or \texttt{EPOCHS} in \passthrough{\lstinline!train.py!} therefore cannot change the result. Arms that kept proposing such edits (``stuck on \texttt{BATCH\_SIZE}'' and ``stuck on \texttt{TRACE\_LEN}'' in Table~\ref{tab:supp-memory-ablation}) repeated edits without effect, which raises their repeated-bad rate, so the replicate comparison partly reflects this interface rather than the memory mode. A small cross-node \texttt{lr\allowbreak\_synthetic} diagnostic (10 trials per memory mode) completes 30/30 valid LangGraph trials but does not reproduce the hypothesised RBR ordering: rationale (0.80) exceeds both none and summary (0.70), the reverse of the hypothesis. One possible reading of this reversal is an anchoring effect --- richer rationale context gives the manager more prior failed patterns to fixate on rather than redirect from; either way, the result supports treating repeated-bad rate as a per-node diagnostic rather than as evidence of a general memory-mode effect.

\textbf{Verbosity ablation.} In this single run, truncating rationale to 50 tokens does not restore summary-level performance (RBR(C) = 0.80 = RBR(D) $\neq$ RBR(B) = 0.50), so rationale \emph{length} alone does not explain the difference. The full-rationale arm of this run was stuck on \texttt{BATCH\_SIZE}, an edit without effect in the fast-search mode, and the truncated arm ended at the same starting AUC, so this comparison is weak evidence.

\subsection{Large-Scale Memory Replication (NVIDIA L40S, 225 Trials)}\label{large-scale-replication}

The large-scale L40S replication (design and models in Section~\ref{memory-ablation-design}: 3 memory modes $\times$ 5 independent seeds $\times$ 15 trials = 225 trials) tests the full hypothesised ordering with variance estimates. Each seed reset \passthrough{\lstinline!train.py!} to a verified clean baseline before its first trial; reset integrity is confirmed by a shared \texttt{node\allowbreak\_state\allowbreak\_hash} at trial~1 across all 15 campaigns.

\textbf{Starting point and configuration.} All arms start from the same level: trial-1 \texttt{val\_auc} lies between 0.886 and 0.890 in every arm and seed (Table~\ref{tab:supp-resnet-linux}, Fig.~\ref{fig:supp-trajectory}). This level is well above the case study's initial 0.7747, and the configurations differ: the L40S campaigns train \texttt{resnet\_trigger} in its full mode on the GPU (4{,}000 signal and 4{,}000 noise traces of 8{,}000 samples; batch size 32 and 5 epochs before any edit), starting from the \passthrough{\lstinline!train.py!} committed on 30 April 2026 (harness commit \texttt{d589d88}), which reproduces the shared trial-1 \texttt{node\allowbreak\_state\allowbreak\_hash}, whereas the case study, the LangGraph ablation and the baseline-seed interval use the node's CPU fast-search mode (1{,}000 signal and 1{,}000 noise traces cropped to 4{,}096 samples; batch size 64 and 3 epochs, fixed by the mode). L40S AUCs are therefore comparable across arms but not with the other ResNet results. From the common start, the mean best AUC rises by about 0.011 without memory and by about 0.05 with either memory mode.

\textbf{L40S replication: memory arms outperform none.} Across all five clean seeds, mean best AUC follows none $<$ summary $\approx$ rationale (0.899 vs.\ 0.939 and 0.941); the 0.002 difference between the two memory arms is not treated as meaningful. Cross-seed variability is lower for both memory arms, with $\sigma_{\text{none}} = 0.009$ vs.\ $\sigma_{\text{summary}} = \sigma_{\text{rationale}} = 0.002$, a 4--5$\times$ smaller standard deviation; with five seeds per arm this is descriptive. Memory arms diverge from the no-memory arm by trial~5 and hold the gap through trial~15 (Figs.~\ref{fig:supp-trajectory} and~\ref{fig:supp-resnet-linux-panels}). In an earlier run of these 15 campaigns the reset of \passthrough{\lstinline!train.py!} failed: within seeds s1--s3 each arm started from a different file, and s4--s5 started from a modified baseline. Those ledgers were discarded (deleted by the rerun scripts, outside the control plane), and all 15 campaigns were rerun under the same campaign IDs from a verified clean baseline; the numbers reported here come from the rerun. Provenance completeness: 100\% across all 225 trials; the memory effect remains node- and manager-specific.

\subsection{Real LM Training Node: Governance Under High Failure Rate}\label{autoresearch-linux}

The \texttt{autoresearch\allowbreak\_linux} node (GPT pretraining on the autoresearch data shards, NVIDIA L40S, metric \texttt{val\_bpb}) is reported as a governance stress case under high failure rate. The \texttt{none} arm produces IR\,=\,1.00 across all 120 trials (four seeds); three of four summary seeds produce valid improving trials (mean \texttt{val\allowbreak\_bpb} reduction $-0.062$), while summary seed~s2 also suffers total worker failure; one of its 30 records was written by the pending-trial recovery for a trial interrupted by a crash (Appendix~\ref{sec:supp-autoresearch-table}). Unlike the corrected ResNet replication, these campaigns were not run with the later frozen-baseline restore protocol, and the archived trial-1 hashes do not verify a common start state across all arms and seeds. We therefore report both the \texttt{none} arm's total failure and summary seed~s2 as observed failures of undetermined cause (an infrastructure or protocol anomaly cannot be excluded), not as evidence for a memory effect. Across 240 trials including 151 runtime-error crashes, provenance completeness is 240/240 (100\%): every \texttt{runtime\allowbreak\_error} event is machine-readable with zero silent failures. The governance finding is classification fidelity under failure, not a general claim about memory-free managers.

\section{Discussion and Limitations}\label{discussion-and-limitations}

\noindent\textbf{Evidence strength.}
Table~\ref{tab:evidence-strength} separates demonstrated properties from case-study findings and non-claims.

\begin{table}[!ht]
\scriptsize
\renewcommand{\arraystretch}{0.72}
\setlength{\tabcolsep}{2pt}
\caption{Evidence strength by claim.}
\label{tab:evidence-strength}
\begin{tabular}{@{}p{0.40\columnwidth}p{0.28\columnwidth}p{0.27\columnwidth}@{}}
\toprule
Claim & Evidence level & Where \\
\midrule
Control plane classifies kept / discarded / failed\_invalid & Demonstrated & Secs~5.1--5.6 \\
Append-only ledger: 100\% provenance & Demonstrated & All nodes; Sec~\ref{cross-node-summary} \\
Guard failures visible in audit record & Demonstrated & Stress, Sec~\ref{stress-campaigns} \\
Protocol transfers across node types & Demonstrated, 6 nodes & Secs~5.2--5.6 \\
Repeated-bad rate diagnoses memory sensitivity & Mixed case study & ResNet: summary lowest in 2/3; \texttt{lr\_synthetic} fails \\
Memory ordering none $<$ summary $\approx$ rationale (AUC) & Observed, 5 clean seeds & Sec~\ref{large-scale-replication}; not on \texttt{lr\_synthetic} \\
Smaller cross-seed spread with memory & Observed, 5 seeds (descriptive) & $\sigma_{\text{none}}$=0.009 vs.\ 0.002 (4--5$\times$) \\
Rationale reduces repeated-bad rate & Not claimed & Non-monotonic; verbosity ablation not supported \\
No-memory arm: 100\% invalid, cause undetermined (start state unverified) & Observed, 4 seeds & Sec~\ref{autoresearch-linux} \\
Governance classifies total worker failure & Demonstrated & Sec~\ref{autoresearch-linux} \\
\bottomrule
\end{tabular}
\end{table}

\noindent\textbf{Limitations.}
Four limitations bound the claims. \emph{Scope of evidence:} six nodes broaden evidence beyond a single case study, but budgets are short, the summary arm has the lowest RBR in only 2 of 3 ResNet replicates (the full hypothesised ordering holds only in rep2, whose arms were not reset between campaigns), and the ordering reverses on \texttt{lr\allowbreak\_synthetic}. \emph{Label validation:} structural relabeling agreement ($\kappa{=}0.830$, Section~\ref{metric-validation-probes}) is a consistency check, not a human study; two-rater validation and holdout-node replication remain future work. \emph{Task metric and reset scope:} the ResNet best AUC falls within the 5-seed baseline min--max range [0.771,\,0.811] (bootstrap 95\% CI of mean: [0.775,\,0.798]), so AUC gain is secondary evidence; the \texttt{autoresearch\allowbreak\_linux} campaigns predate the frozen-baseline restore protocol, so neither their \texttt{none}-arm total failure nor their s2 anomaly is used as memory-effect evidence. \emph{Human utility:} the paper shows that governance properties can be enforced and measured, and what evidence is lost without them (Section~\ref{stress-campaigns}) --- not that governed ledgers improve human audit outcomes; a human-auditor utility study (do reviewers with governed ledgers detect seeded errors faster?) is the natural next step.

\section{Conclusion}
\label{sec:conclusion}

Across 1{,}445 governed trials on six ML nodes, the control plane classifies
every outcome without silent corruption; provenance is complete by
construction and is observed in all 1{,}445 trials, as the invariant
requires. Autonomous ML experimentation should
be evaluated on governance metrics---provenance, lifecycle classification,
scope enforcement, and failure taxonomy---not only task metrics.

\begin{acks}
The author is deeply grateful to Sudhir Shetty for the many inspiring
discussions, and for his generous support and patience along the way. The
author also warmly thanks Tony Chen and Girish Hoogar for their encouragement
and for the thoughtful conversations that have enriched his perspective and
helped shape his research interests.
\end{acks}

\section*{AI Usage Statement}
Apart from the LLMs studied here, the author used AI assistants, including
Anthropic's Claude, to help write the harness code and to edit and check this
manuscript, and reviewed all of their output.

\clearpage
\bibliographystyle{ACM-Reference-Format}
\bibliography{references}

\clearpage
\appendix

\setcounter{figure}{0}
\setcounter{table}{0}
\renewcommand{\thefigure}{S\arabic{figure}}
\renewcommand{\thetable}{S\arabic{table}}
\renewcommand{\theHfigure}{S\arabic{figure}}
\renewcommand{\theHtable}{S\arabic{table}}

These appendices are the reproducibility material of the paper (the
supplement of the workshop version). They provide the ledger schema, NodeSpec
contract examples, control-plane pseudocode, extended experiment tables, the
failure taxonomy with concrete ledger examples, campaign identifiers, the
hardware and model environment, and notebook details. No central claim in the
main text depends only on these appendices.

\section{Reproducibility and Artifact Description}
\label{sec:supp-reproducibility}

\subsection{Artifact and Code Availability}
\label{sec:supp-artifact}

{\sloppy The harness code is public at
\url{https://github.com/dowlingwong/autoresearch_harness} under the MIT
licence. The trial ledgers behind the numbers in this paper are released as a
tagged GitHub release of that repository, archived on Zenodo:
release \texttt{kdd-etaai-arxiv-v1} (commit \texttt{0455171}), DOI~\href{https://doi.org/10.5281/zenodo.24174991}{10.5281/zenodo.24174991}. The release holds the 60
primary ledgers of Table~\ref{tab:supp-campaign-inventory}, the validation
and counterfactual ledgers of Table~\ref{tab:supp-validation-campaigns} (the
ungoverned \texttt{autoresearch\allowbreak\_linux} arm recorded no valid trial,
so it has no ledger), the
two \texttt{*\_ungoverned\_obs.jsonl} observation logs, and an
\texttt{artifact\_manifest.json} that lists each file with its SHA-256 digest
and trial count. Each released ledger and observation log is byte-identical to
its copy in \path{autoresearch-harness-artifact.zip}, which was committed to
the same repository on 25 May 2026 (commit \texttt{62e93a8}) and remains in
its history. The tagged commit is the code version from which the
reported numbers are recomputed; it supersedes commit \texttt{ff3717b}, which
the workshop version cited and which does not contain the ledgers. The ledgers of
the ResNet-trigger LangGraph memory ablation (all rows of
Table~\ref{tab:supp-memory-ablation} except the \texttt{lr\_synthetic} rows) are
not in the tagged release; they are in the same archive of commit
\texttt{62e93a8}, from which those rows are recomputed.\par}

{\sloppy The release archive contains: \path{src/} (control-plane source),
\path{configs/nodes/} (node YAML specs), \path{scripts/} (campaign runners),
\path{tests/} (test suite), \path{experiments/ledgers/} (the ledgers
listed above), \path{nodes/} (per-node training code), \path{paper_figures.ipynb}
(figure notebook), \path{artifact_manifest.json} (ledger index), and
\path{ARTIFACT_README.md} (step-by-step instructions).
It excludes large or private execution artifacts: the two ResNet HDF5 data
files and the execution-host patch and run-log files are not needed to
recompute the reported governance metrics from the ledgers.\par}

Execution-host patch and run-log files are referenced in each ledger record by
path fields and provenance IDs; the files themselves are not part of the
release.

Each trial record contains a trial identifier, campaign identifier,
proposal summary/rationale, patch reference, raw-log reference, parsed metrics,
terminal decision, failure category where applicable, node-state hash, and
provenance IDs for proposal, patch, run, metric, and decision objects. Readers
can map each paper table to its ledgers under
\texttt{experiments/\allowbreak ledgers/} and from there to the artifact
references recorded in each ledger row.

\subsection{Ledger Schema}
\label{sec:supp-ledger-schema}

\begin{table}[ht]
\centering
\footnotesize
\caption{Core ledger fields. Terminal lifecycle states are \texttt{kept},
\texttt{discarded}, and \texttt{failed\_invalid}; failure labels are attached
only where a control-plane guard or worker failure produces a failure condition.}
\label{tab:supp-ledger-schema}
\begin{tabular}{@{}p{0.30\linewidth}p{0.62\linewidth}@{}}
\toprule
Field & Meaning \\
\midrule
\texttt{trial\_id} & Stable identifier for one budget slot. \\
\texttt{campaign\_id} & Links rows to the campaign ID inventory in
  Section~\ref{sec:supp-campaign-ids}. \\
\texttt{budget\_index} & Trial index within the fixed campaign budget. \\
\texttt{decision} & Terminal state: \texttt{kept}, \texttt{discarded}, or
  \texttt{failed\_invalid}. \\
\texttt{failure\_category} & Control-plane label such as \texttt{runtime\_error},
  \texttt{invalid\_edit\_scope}, \texttt{no\_op\_patch},
  \texttt{proposal\_precondition\_failed}, \texttt{invalid\_config},
  \texttt{metric\_missing}, or \texttt{edit\_failed}; valid-but-worse trials use
  \texttt{degraded\_metric}
  and are \texttt{discarded}. \\
\texttt{proposal\_summary}, \texttt{proposal\_rationale} & Manager proposal text
  recorded before worker execution. \\
\texttt{patch\_ref}, \texttt{patch\_hash} & Patch artifact path and content hash
  when a patch is materialised. \\
\texttt{raw\_log\_ref} & Worker run-log artifact path. \\
\texttt{parsed\_metrics} & Machine-parsed task metrics used by the decision rule. \\
\texttt{provenance} & Stable IDs for the trial's proposal, patch, run, metric,
  and decision objects, each derived by hashing the trial ID. \\
\texttt{node\_state\_hash} & Hash of node-local state used for reset and
  divergence auditing. \\
\bottomrule
\end{tabular}
\end{table}

\subsection{Dry-Run Commands}

The harness can be exercised without Ollama or a GPU using dry-run mode:

\begin{lstlisting}
# Smoke test --- 3 trials, synthetic metrics, no LLM or node required
python3 scripts/run_campaign.py \
    --node resnet_trigger --campaign-id smoke \
    --budget 3 --dry-run

# LangGraph manager smoke test (stub LLM, no Ollama)
python3 scripts/run_campaign.py \
    --node resnet_trigger --campaign-id lg_smoke \
    --budget 2 --manager langgraph_manager \
    --dry-run --llm-stub

# Memory ablation dry run (all three modes)
python3 scripts/run_memory_ablation.py \
    --node resnet_trigger --budget 3 --dry-run
\end{lstlisting}

For a real campaign with a local Ollama model:
\begin{lstlisting}
python3 scripts/run_campaign.py \
    --node resnet_trigger --campaign-id main \
    --budget 5 --manager prompt_manager \
    --memory-mode append_only_summary_with_rationale \
    --node-root nodes/ResNet_trigger \
    --model qwen2.5-coder:7b \
    --host http://localhost:11434
\end{lstlisting}

\subsection{Hardware and Model Environment}

{\sloppy Large-scale ResNet and \texttt{autoresearch\_linux} campaigns ran on
an institutional GPU cluster (NVIDIA L40S, CUDA 12.8, Ubuntu 22.04,
PyTorch 2.x). Cross-node LangGraph campaigns use
\texttt{deepseek/\allowbreak deepseek-v4-flash}, the Flash model of the DeepSeek V4
preview released on 24 April 2026~\cite{deepseek2026v4}, through the
OpenAI-compatible DeepSeek API
endpoint (\texttt{DEEPSEEK\allowbreak\_BASE\allowbreak\_URL},
default \url{https://api.deepseek.com};
authenticated via \texttt{DEEPSEEK\allowbreak\_API\allowbreak\_KEY}).
By their ledger timestamps (UTC), all campaigns in
Tables~\ref{tab:supp-campaign-inventory} and~\ref{tab:supp-validation-campaigns}
ran between 9 and 22 May 2026; the corrected L40S rerun ran on 21 May and the
counterfactual campaigns on 22 May.
The ResNet L40S replication uses \texttt{qwen2.5-coder:7b} at temperature 0.2.
The stochastic memory ablation uses \texttt{qwen2.5-coder:7b} at temperature 0.7
with the explicit avoidance prompt and deterministic patch bridge. LocalWorker
campaigns run under Python 3.11 on a CPU-only host.\par}

Hosted API proposals may vary if the provider changes the served model behind a
fixed identifier. Exact proposal text is most reproducible from the released
ledgers. Governance metrics remain recomputable from archived trial
records without API access. No API keys, account identifiers, or private endpoint
secrets are included in the artifact.

\subsection{Bootstrap Confidence Intervals}

Seed-level bootstrap confidence intervals use 10{,}000 resamples with random
seed 42. Resampling is performed over seed-level best metrics, not individual
trial records, so each interval reflects between-seed variability.

\subsection{Node-State-Hash Reset Verification}
\label{sec:supp-reset-hash}

Each trial record contains a \texttt{node\_state\_hash} computed from the SHA-256
digest of the node's editable file set immediately before the worker executes.
This field allows independent verification that every seed of a multi-seed ablation
campaign started from an identical node state --- a necessary condition for fair
memory-mode comparisons. For the L40S replication
(Table~\ref{tab:supp-resnet-linux}), all five seeds
produce hash prefix \texttt{9c0118f33fcb} at \texttt{budget\_index=1}:

\begin{lstlisting}
python3 -c "
import json
for l in open('experiments/ledgers/<campaign_id>_trials.jsonl'):
    r = json.loads(l)
    if r['budget_index'] == 1:
        print(r['trial_id'], r['node_state_hash'][:12])
"
\end{lstlisting}

A divergence in the prefix across seeds indicates a reset failure and invalidates
the ablation comparison for that seed. The hash is stored in every ledger record,
so the check is replayable from the released ledgers without re-running any
training code.

\subsection{Provenance and Artifact Completeness}
\label{sec:supp-fig-provenance}

\begin{figure}[ht]
\centering
\includegraphics[width=\linewidth]{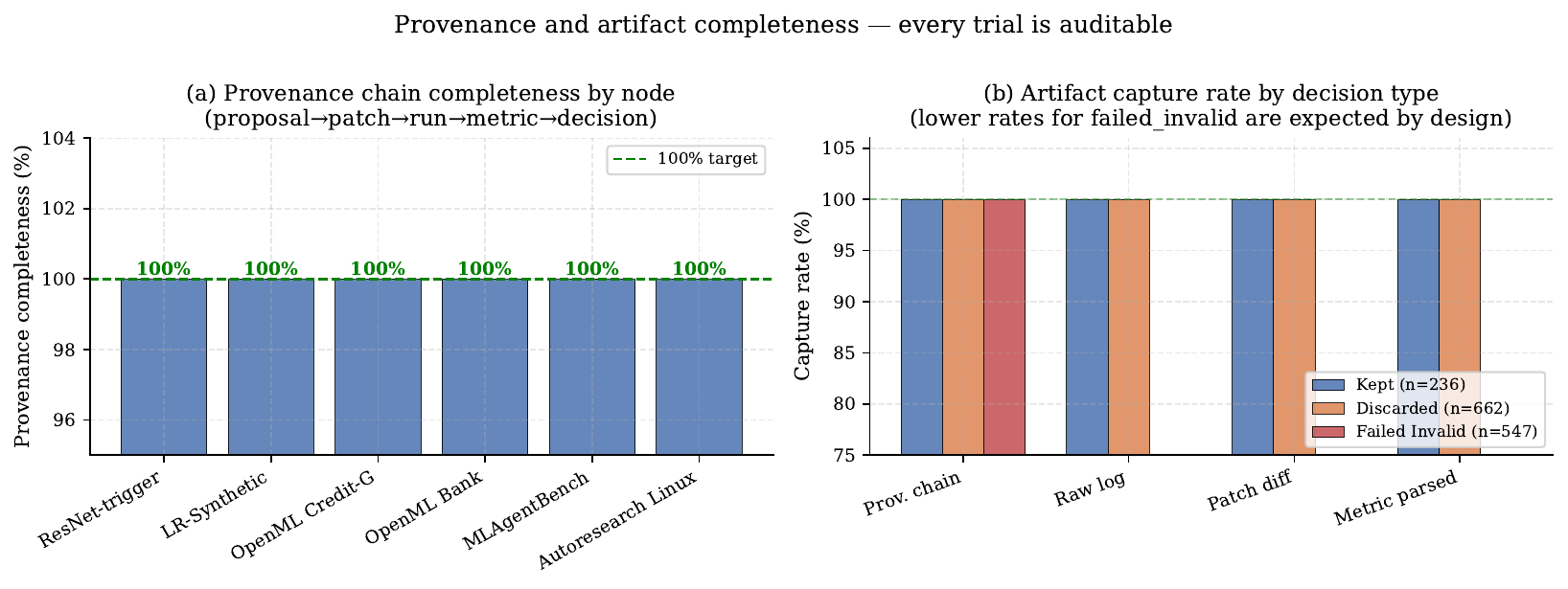}
\Description{Per-node provenance completeness and artifact-reference capture rates by decision type across all governed trials.}
\caption{Provenance and artifact-reference completeness across all 1{,}445 governed trials.
(a) Provenance chain completeness is 100\% on every node. (b) Artifact-reference capture
rates by decision type: lower patch and metric rates for \texttt{failed\_invalid}
are expected (pre-run failures produce no patch; runtime failures produce no
parsed metric).}
\label{fig:supp-provenance}
\end{figure}

Figure~\ref{fig:supp-provenance} shows per-node provenance completeness and
artifact-reference capture rates by decision type. Panel~(a) confirms 100\% provenance
completeness on all six nodes. Panel~(b) shows that lower artifact-reference capture rates
for \texttt{failed\allowbreak\_invalid} trials are expected by design: trials that
fail at the precondition stage (before worker execution) produce no patch diff,
and trials that fail at runtime may produce no parsed metric. The provenance chain
itself (proposal~$\to$ patch~$\to$ run~$\to$ metric~$\to$ decision IDs) is always
complete regardless of the terminal outcome.

\subsection{Analysis Notebook}

{\sloppy The release includes \texttt{paper\_figures.ipynb}. The notebook
loads the campaign ledgers from \texttt{experiments/\allowbreak ledgers/},
treats the append-only ledgers as the source of truth, and writes derived
outputs to \texttt{paper/\allowbreak tables/} and
\texttt{paper/\allowbreak figures/}. Install dependencies and run:\par}
\begin{lstlisting}
uv pip install -e ".[dev,notebook]"
# then: Kernel -> Restart & Run All
\end{lstlisting} The notebook is an analysis convenience
layer; the ledgers, not rendered plots, are the evidence source.

Selected notebook outputs (after final L40S data synchronisation):
{\raggedright\texttt{fig1\allowbreak\_governance\allowbreak\_overview},
\texttt{fig3\allowbreak\_ir\allowbreak\_taxonomy},
\texttt{fig5\allowbreak\_cross\allowbreak\_node},
\texttt{fig6\allowbreak\_provenance},
\texttt{fig7\allowbreak\_openml}, and
\texttt{fig9\allowbreak\_resnet\allowbreak\_linux\allowbreak\_ablation}.\par}
Retired or exploratory outputs (including the MLP synthetic diagnostic plot)
should not be cited as reported evidence unless the main text explicitly
restores that node to the evidence set.

\section{Extended Experiment Tables}
\label{sec:supp-tables}
\subsection{Campaign ID Inventory}
\label{sec:supp-campaign-ids}

Table~\ref{tab:supp-campaign-inventory} lists the 60 primary campaign ledgers
behind the 1{,}445-trial evidence set; Table~\ref{tab:supp-validation-campaigns}
lists the validation and counterfactual campaigns reported outside that total.

\begin{table*}[t]
\centering
\footnotesize
\caption{Primary campaign identifiers for the 1{,}445-trial governed evidence
set. One ledger file exists per listed seed. Campaign IDs with ranges
(e.g.\ \texttt{s1}--\texttt{s5}) denote five separate JSONL ledgers.}
\label{tab:supp-campaign-inventory}
\begin{tabular}{@{}p{0.38\linewidth}p{0.17\linewidth}p{0.17\linewidth}p{0.08\linewidth}p{0.08\linewidth}@{}}
\toprule
Campaign ID(s) & Node & Arm / mode & Seed(s) & Budget \\
\midrule
\texttt{kdd\_resnet\_scientific\_20} & \texttt{resnet\_trigger} & case study & -- & 20 \\
\texttt{deepseek\allowbreak\_resnet\allowbreak\_none\allowbreak\_s1}--\texttt{s5} & \texttt{resnet\allowbreak\_trigger} & none & s1--s5 & 15 \\
\texttt{deepseek\allowbreak\_resnet\allowbreak\_append\allowbreak\_only\allowbreak\_summary\allowbreak\_s1}--\texttt{s5} & \texttt{resnet\allowbreak\_trigger} & summary & s1--s5 & 15 \\
\texttt{deepseek\allowbreak\_resnet\allowbreak\_append\allowbreak\_only\allowbreak\_summary\allowbreak\_with\allowbreak\_rationale\allowbreak\_s1}--\texttt{s5} & \texttt{resnet\allowbreak\_trigger} & rationale & s1--s5 & 15 \\
\texttt{deepseek\allowbreak\_lr\allowbreak\_none\allowbreak\_b30\allowbreak\_s1}--\texttt{s5} & \texttt{lr\allowbreak\_synthetic} & none & s1--s5 & 30 \\
\texttt{deepseek\allowbreak\_lr\allowbreak\_append\allowbreak\_only\allowbreak\_summary\allowbreak\_b30\allowbreak\_s1}--\texttt{s5} & \texttt{lr\allowbreak\_synthetic} & summary & s1--s5 & 30 \\
\texttt{deepseek\allowbreak\_lr\allowbreak\_append\allowbreak\_only\allowbreak\_summary\allowbreak\_with\allowbreak\_rationale\allowbreak\_b30\allowbreak\_s1}--\texttt{s5} & \texttt{lr\allowbreak\_synthetic} & rationale & s1--s5 & 30 \\
\texttt{deepseek\allowbreak\_openml\allowbreak\_cg\allowbreak\_s1}--\texttt{s4} & \texttt{openml\allowbreak\_credit\allowbreak\_g} & summary & s1--s4 & 20 \\
\texttt{deepseek\allowbreak\_openml\allowbreak\_cg\allowbreak\_b30\allowbreak\_s1}--\texttt{s5} & \texttt{openml\allowbreak\_credit\allowbreak\_g} & summary & s1--s5 & 30 \\
\texttt{deepseek\allowbreak\_openml\allowbreak\_bm\allowbreak\_s1}--\texttt{s4} & \texttt{openml\allowbreak\_bank\allowbreak\_marketing} & summary & s1--s4 & 20 \\
\texttt{deepseek\allowbreak\_openml\allowbreak\_bm\allowbreak\_b30\allowbreak\_s1}--\texttt{s5} & \texttt{openml\allowbreak\_bank\allowbreak\_marketing} & summary & s1--s5 & 30 \\
\texttt{mlagentbench\allowbreak\_vectorization\allowbreak\_main\allowbreak\_30} & \texttt{mlagentbench\allowbreak\_vec.} & adapter main & -- & 30 \\
\texttt{deepseek\allowbreak\_mlagentbench\allowbreak\_s1}, \texttt{s2} & \texttt{mlagentbench\allowbreak\_vec.} & adapter rerun & s1--s2 & 10 \\
\texttt{deepseek\allowbreak\_autoresearch\allowbreak\_linux\allowbreak\_none\allowbreak\_s1}--\texttt{s4} & \texttt{autoresearch\allowbreak\_linux} & none & s1--s4 & 30 \\
\texttt{deepseek\allowbreak\_autoresearch\allowbreak\_linux\allowbreak\_append\allowbreak\_only\allowbreak\_summary\allowbreak\_s1}--\texttt{s4} & \texttt{autoresearch\allowbreak\_linux} & summary & s1--s4 & 30 \\
\bottomrule
\end{tabular}
\end{table*}

\begin{table*}[t]
\centering
\footnotesize
\caption{Validation and counterfactual campaign identifiers reported outside the
1{,}445-trial headline total. Ungoverned counterfactual arms append only
valid-trial records to the ledger by design; the full trial denominators are
recorded in the paired \texttt{*\_ungoverned\_obs.jsonl} observation logs
(Table~\ref{tab:counterfactual}).}
\label{tab:supp-validation-campaigns}
\begin{tabular}{@{}p{0.38\linewidth}p{0.17\linewidth}p{0.17\linewidth}p{0.08\linewidth}p{0.08\linewidth}@{}}
\toprule
Campaign ID(s) & Node & Arm / mode & Seed(s) & Budget \\
\midrule
\texttt{kdd\allowbreak\_stress\allowbreak\_scope} & \texttt{resnet\allowbreak\_trigger} & scope stress & -- & 1 \\
\texttt{kdd\allowbreak\_stress\allowbreak\_noop} & \texttt{resnet\allowbreak\_trigger} & no-op stress & -- & 1 \\
\texttt{lr\allowbreak\_synth\allowbreak\_baseline} & \texttt{lr\allowbreak\_synthetic} & baseline manager & -- & 5 \\
\texttt{lr\allowbreak\_synth\allowbreak\_lg\allowbreak\_none} & \texttt{lr\allowbreak\_synthetic} & none & -- & 10 \\
\texttt{lr\allowbreak\_synth\allowbreak\_lg\allowbreak\_summary} & \texttt{lr\allowbreak\_synthetic} & summary & -- & 10 \\
\texttt{lr\allowbreak\_synth\allowbreak\_lg\allowbreak\_rationale} & \texttt{lr\allowbreak\_synthetic} & rationale & -- & 10 \\
\texttt{kdd\allowbreak\_cf\allowbreak\_openml\allowbreak\_gov} & \texttt{openml\allowbreak\_bank\allowbreak\_marketing} & governed (matched) & -- & 20 \\
\texttt{kdd\allowbreak\_cf\allowbreak\_openml\allowbreak\_ung} & \texttt{openml\allowbreak\_bank\allowbreak\_marketing} & ungoverned (matched) & -- & 20 \\
\texttt{kdd\allowbreak\_cf\allowbreak\_arlinux\allowbreak\_gov} & \texttt{autoresearch\allowbreak\_linux} & governed (matched) & -- & 30 \\
\texttt{kdd\allowbreak\_cf\allowbreak\_arlinux\allowbreak\_ung} & \texttt{autoresearch\allowbreak\_linux} & ungoverned (matched) & -- & 30 \\
\bottomrule
\end{tabular}
\end{table*}

\subsection{LangGraph Memory Ablation Detail}
\label{sec:supp-lg-ablation}

Table~\ref{tab:supp-memory-ablation} gives the arm-level results of the
LangGraph memory ablation summarised in Section~\ref{memory-ablation}.

\begin{table*}[t]
\centering
\footnotesize
\caption{LangGraph memory ablation (qwen2.5-coder:7b,
temp=0.7, avoidance prompt, deterministic patch bridge). RBR\,=\,repeated-bad
rate, the repeated-bad count divided by the budget $N$
(Appendix~\ref{sec:supp-rbr}), recomputed for every row from the ledgers. The
workshop version reported, for the ResNet-trigger rows, the count over the first
$N{-}1$ trials divided by $N{-}1$ (e.g., 0.78\,=\,7/9); the order of the arms
within each replicate is the same under both. The rep2 arms were not reset
between campaigns (Section~\ref{memory-ablation}). Arm C tests whether rationale \emph{length} drives Arm D's result. The
\texttt{lr\_synthetic} rows use validation score, not AUC.}
\label{tab:supp-memory-ablation}
\begin{tabular}{@{}p{0.05\linewidth}p{0.22\linewidth}rrrrp{0.35\linewidth}@{}}
\toprule
Arm & Memory mode & RBR & Kept & Disc. & Best metric & Observation \\
\midrule
\multicolumn{7}{l}{\textit{Primary ablation --- rep1, budget=10}} \\
A & none & 0.80 & 1 & 9 & 0.7747 & Baseline \\
B & summary & 0.50 & 4 & 6 & 0.8212 & Discovers \texttt{TRAIN\_FRACTION} \\
D & full rationale & 0.80 & 1 & 9 & 0.7747 & Stuck on \texttt{BATCH\_SIZE} \\
\midrule
\multicolumn{7}{l}{\textit{Verbosity ablation --- same node/budget/model}} \\
C & short rationale (50\,tok) & 0.80 & 1 & 9 & 0.7747 & Matches D, not B: length alone does not explain D \\
\midrule
\multicolumn{7}{l}{\textit{Seed replication --- rep2, budget=10; arms not reset between campaigns}} \\
A & none & 0.80 & 1 & 9 & 0.7747 & \\
B & summary & 0.70 & 2 & 8 & 0.7903 & Summary not lowest in this replicate \\
D & full rationale & 0.60 & 3 & 7 & 0.8220 & Lowest RBR in rep2; started from B's final state \\
\midrule
\multicolumn{7}{l}{\textit{Budget replication --- rep3, budget=20}} \\
A & none & 0.85 & 2 & 18 & 0.8031 & \\
B & summary & 0.80 & 3 & 17 & 0.8150 & Summary lowest; finds \texttt{TRAIN\_FRACTION} at T5 \\
D & full rationale & 0.85 & 1 & 17 & 0.7747 & Stuck on \texttt{TRACE\_LEN}; also 2 failed-invalid (\texttt{no\_op\_patch}) \\
\midrule
\multicolumn{7}{l}{\emph{Cross-node transfer} --- \texttt{lr\_synthetic}, budget=10} \\
A & none & 0.70 & 2 & 8 & 0.9230 & Clean LocalWorker run \\
B & summary & 0.70 & 2 & 8 & 0.9228 & No RBR reduction on this node \\
D & full rationale & 0.80 & 1 & 9 & 0.9227 & Slightly higher RBR \\
\bottomrule
\end{tabular}
\end{table*}

\clearpage
\subsection{Cross-Node Acceptance and Invalid Rates}
\label{sec:supp-fig-crossnode}

\begin{figure}[ht]
\centering
\includegraphics[width=\linewidth]{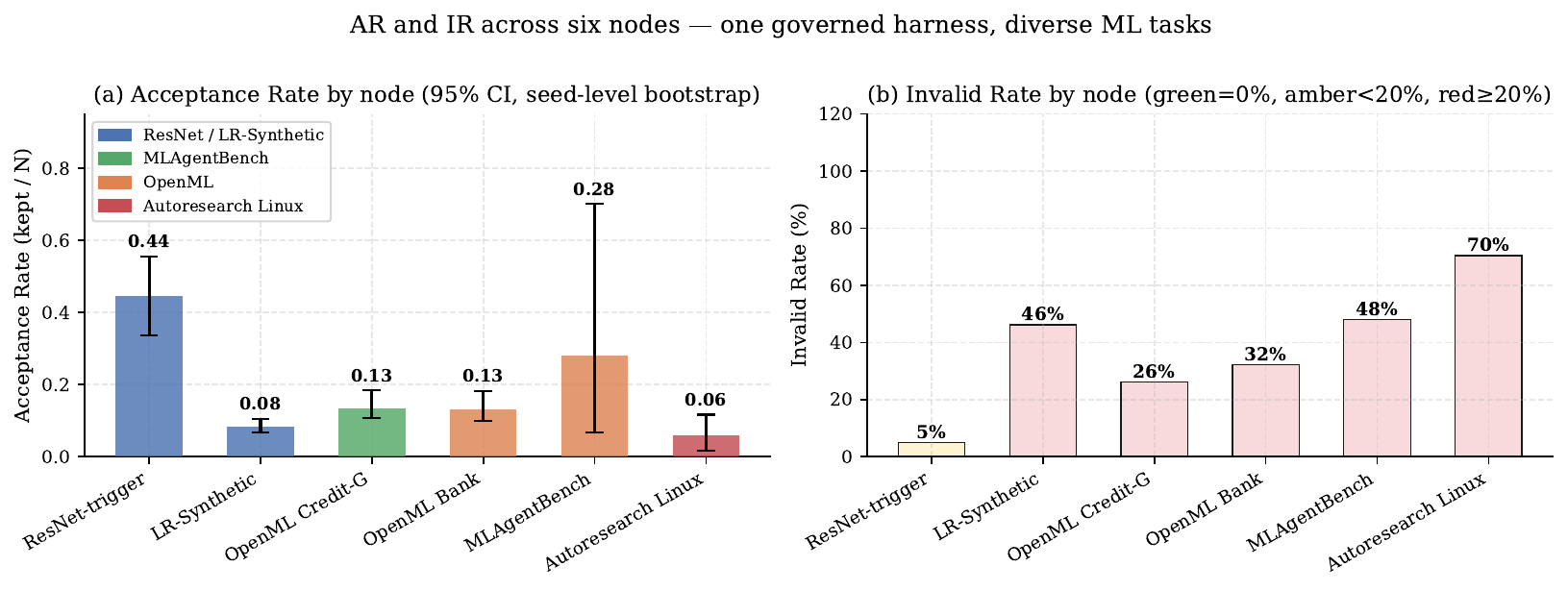}
\Description{Cross-node acceptance and invalid rates with seed-level bootstrap confidence intervals.}
\caption{Cross-node acceptance rate (AR) and invalid rate (IR). Bootstrap CIs
use 10{,}000 seed-level resamples. Colour denotes task family. AR and IR reflect
proposal quality and task difficulty, not governance failures.}
\label{fig:supp-crossnode}
\end{figure}

Figure~\ref{fig:supp-crossnode} summarises acceptance rate (AR) and invalid rate
(IR) across all six nodes with 95\% seed-level bootstrap confidence intervals.
AR and IR vary substantially by node and manager interface. The governance claim
is not that these rates are uniform, but that the lifecycle classification and
provenance record are correct on every node regardless of the rates observed.

\subsection{LR-Synthetic Memory Ablation}
\label{sec:supp-fig-lr}

\begin{figure}[ht]
\centering
\includegraphics[width=\linewidth]{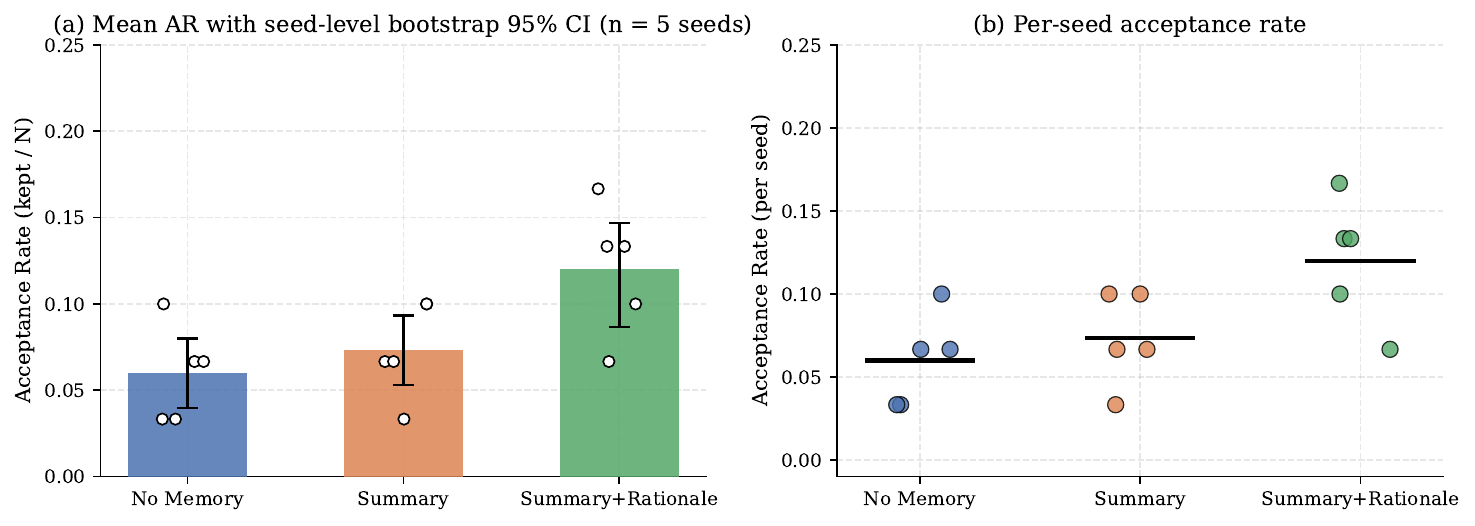}
\Description{LR-synthetic memory ablation with mean acceptance rates and per-seed scatter for none, summary, and rationale memory modes.}
\caption{LR-synthetic memory ablation (150 trials per arm: 5 seeds of 30
trials). Left: mean
AR with seed-level bootstrap 95\% CIs; right: per-seed AR. The rationale arm's
interval lies above the no-memory arm's (descriptive; five seeds per arm).}
\label{fig:supp-lr}
\end{figure}

Figure~\ref{fig:supp-lr} shows the memory ablation on \texttt{lr\_synthetic}
(450 trials: three arms, five seeds per arm, 30 trials per seed;
Table~\ref{tab:supp-campaign-inventory}). Acceptance is low in every arm. The rationale
arm has the highest mean acceptance rate, and its seed-level bootstrap interval
lies above that of the no-memory arm; with five seeds per arm this difference
is descriptive and untested. The control plane classifies every trial in all
three arms, so governance transfers to this node whatever its memory response,
which differs from the ResNet response (Section~\ref{large-scale-replication}).

\subsection{ResNet L40S Best-AUC Trajectory}
\label{sec:supp-fig-trajectory}

\begin{figure}[t]
\centering
\includegraphics[width=\linewidth]{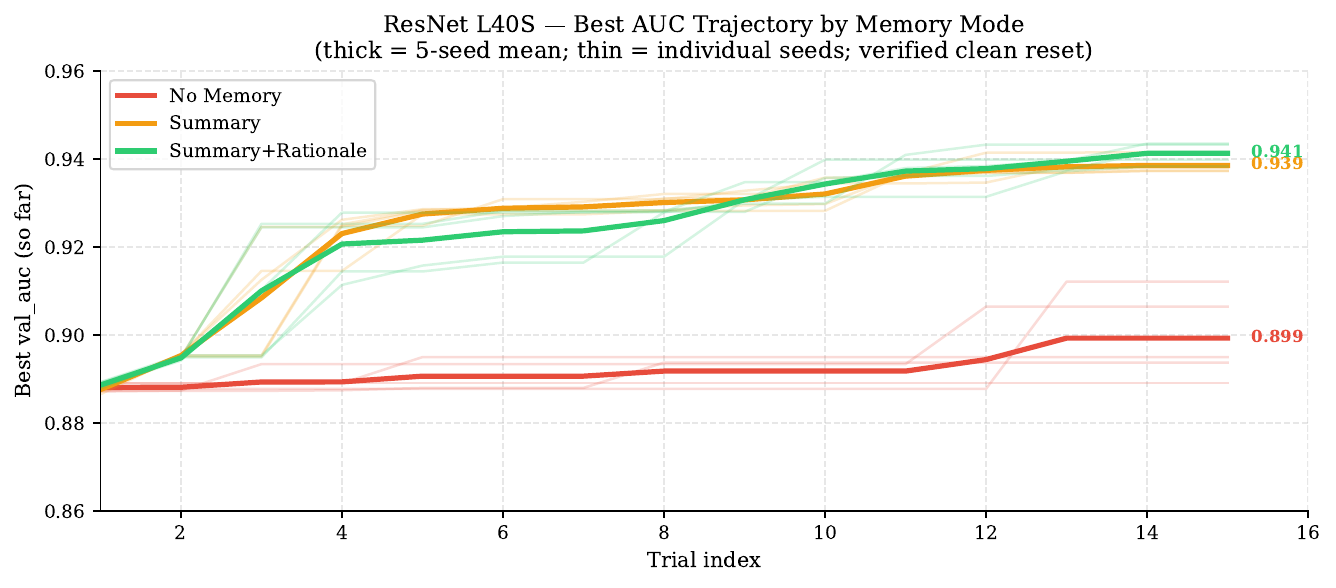}
\Description{Best-AUC trajectory by memory mode across 15 trials on ResNet L40S, with thick lines showing 5-seed means and thin lines showing individual seeds for none, summary, and rationale modes.}
\caption{ResNet L40S best-AUC trajectory by memory mode (3 modes $\times$ 5 seeds, verified reset). Thick: 5-seed mean; thin: individual seeds. Memory arms diverge from the no-memory arm by trial~5 and hold the gap through trial~15, consistent with the 4--5$\times$ smaller cross-seed standard deviation reported in Section~\ref{large-scale-replication}.}
\label{fig:supp-trajectory}
\end{figure}

Figure~\ref{fig:supp-trajectory} shows the best-AUC trajectory of each arm and
seed of the L40S replication (Section~\ref{large-scale-replication}).

\subsection{Large-Scale L40S Replication}
\label{sec:supp-l40s-table}

Table~\ref{tab:supp-resnet-linux} summarises the L40S replication by memory
mode.

\begin{table*}[t]
\centering
\footnotesize
\caption{Large-scale memory ablation (ResNet-trigger, NVIDIA
L40S, 75 trials per arm = 3\,modes $\times$ 5\,seeds $\times$ 15\,trials).
Reset integrity confirmed by shared \texttt{node\_state\_hash}
(\texttt{9c0118f33fcb}) at trial~1 across all 15 campaigns. Trial-1
\texttt{val\_auc} is given per seed, s1 to s5, from the first record of each
campaign's ledger; it is not comparable with the
case-study values, which use the CPU fast-search mode
(Section~\ref{large-scale-replication}). $\sigma$
is per-seed best-AUC standard deviation. 95\% CIs use 10{,}000
seed-level bootstrap resamples (seed 42). Manager: \texttt{deepseek-v4-flash};
worker: \texttt{qwen2.5-coder:7b} at temperature 0.2.}
\label{tab:supp-resnet-linux}
\begin{tabular}{@{}lrrrrlrrr@{}}
\toprule
Memory mode & Seeds & $N$ & Acc\% & Inv\% & Trial-1 AUC (s1, s2, s3, s4, s5) & Mean best AUC & 95\% CI & $\sigma$ \\
\midrule
none      & 5 & 75 & 17.3 &  0.0 & 0.8878, 0.8891, 0.8890, 0.8872, 0.8873 & 0.899 & [0.892,\,0.907] & 0.009 \\
summary   & 5 & 75 & 64.0 &  1.3 & 0.8885, 0.8878, 0.8877, 0.8865, 0.8874 & 0.939 & [0.937,\,0.940] & 0.002 \\
rationale & 5 & 75 & 58.7 &  0.0 & 0.8885, 0.8878, 0.8883, 0.8892, 0.8893 & 0.941 & [0.940,\,0.943] & 0.002 \\
\bottomrule
\end{tabular}
\end{table*}

\subsection{Autoresearch Linux Per-Seed Results}
\label{sec:supp-autoresearch-table}

Table~\ref{tab:supp-autoresearch-linux} gives the per-seed outcomes of the
\texttt{autoresearch\_linux} campaigns (Section~\ref{autoresearch-linux}).
The ledger of summary seed s2 is the only primary ledger with irregular
budget indices: index~1 has no record, and index~2 has two records with the
same trial ID, one for each time the slot was opened. The first was appended
by the pending-trial recovery after an earlier run crashed while that trial
was pending; the second records the trial executed after the restart. All 30
records are counted, so one of the 1{,}445 primary records was written by the
recovery tool rather than at the end of a completed trial.

\begin{table*}[t]
\centering
\footnotesize
\caption{Autoresearch Linux node (NVIDIA L40S,
\texttt{deepseek-v4-flash} manager, \texttt{claw\_style\_worker},
4\,seeds $\times$ 30\,trials per arm). AR\,=\,acceptance rate;
IR\,=\,invalid rate; Gain\,=\,best $-$ initial \texttt{val\_bpb}
(negative = improvement). Seed s2 of the summary arm suffered total
worker failure; task-metric gain summaries use s1/s3/s4. AR uses the definition
of Table~\ref{tab:governance-metrics}, $|\mathrm{kept}|/N$.}
\label{tab:supp-autoresearch-linux}
\begin{tabular}{@{}llrrrrrr@{}}
\toprule
Arm & Seed & Kept & Disc & Fail & AR & IR & Gain \\
\midrule
\multirow{4}{*}{\texttt{none}} & s1 & 0 & 0 & 30 & 0.00 & 1.00 & --- \\
 & s2 & 0 & 0 & 30 & 0.00 & 1.00 & --- \\
 & s3 & 0 & 0 & 30 & 0.00 & 1.00 & --- \\
 & s4 & 0 & 0 & 30 & 0.00 & 1.00 & --- \\
\midrule
\multirow{4}{*}{\texttt{summary}} & s1 & 4 & 17 &  9 & 0.13 & 0.30 & $-$0.0616 \\
 & s2$^{\dagger}$ & 0 & 0 & 30 & 0.00 & 1.00 & --- \\
 & s3 & 5 & 18 &  7 & 0.17 & 0.23 & $-$0.0607 \\
 & s4 & 5 & 22 &  3 & 0.17 & 0.10 & $-$0.0642 \\
\bottomrule
\end{tabular}

\footnotesize $^{\dagger}$ s2 suffered total worker failure (27
\texttt{runtime\_error} + 3 \texttt{proposal\_precondition\_failed}); no
initial \texttt{val\_bpb} was recorded. These campaigns predate the
frozen-baseline restore protocol, and the archived trial-1 hashes do not verify
a common start state across all arms and seeds; s2, like the total failure
of the \texttt{none} arm, is therefore an observed failure of undetermined
cause, not memory-effect evidence. Governance correctly classified all 30
trials as \texttt{failed\_invalid}.
\end{table*}

\subsection{OpenML Per-Budget Summary}
\label{sec:supp-openml-summary}

Table~\ref{tab:supp-openml-campaign-summary} summarises the OpenML campaigns by
node and budget.

\begin{table*}[t]
\centering
\footnotesize
\caption{OpenML campaign summary by public node and budget. Seed-level bootstrap
confidence intervals use 10{,}000 resamples (seed 42).}
\label{tab:supp-openml-campaign-summary}
\begin{tabular}{@{}llrrrrrr@{}}
\toprule
Node & Budget & Seeds & Trials & AR\% & IR\% & Mean best AUC & 95\% CI \\
\midrule
\multirow{2}{*}{\texttt{credit\_g}}    & b20 & 4 &  80 & 20.0 &  8.8 & 0.7671 & [0.7655, 0.7687] \\
                                        & b30 & 5 & 150 & 10.0 & 35.3 & 0.7679 & [0.7667, 0.7688] \\
\multirow{2}{*}{\texttt{bank\_mkt}}    & b20 & 4 &  80 & 20.0 & 17.5 & 0.9295 & [0.9263, 0.9330] \\
                                        & b30 & 5 & 150 &  9.3 & 40.0 & 0.9341 & [0.9337, 0.9345] \\
\bottomrule
\end{tabular}

\end{table*}

\subsection{OpenML Dataset Statistics}
\label{sec:supp-openml-datasets}

\begin{table}[ht]
\centering
\footnotesize
\caption{OpenML dataset characteristics for the two public classification nodes
used in Section~\ref{results}.}
\label{tab:supp-openml-datasets}
\begin{tabular}{@{}llrrlr@{}}
\toprule
ID & Name & $N$ & Feats. & Task & Maj.\ cls. \\
\midrule
31   & credit-g       & 1{,}000  & 20 & Binary & 70\% \\
1461 & bank-marketing & 45{,}211 & 16 & Binary & 88\% \\
\bottomrule
\end{tabular}
\end{table}

\noindent The class imbalance in the bank-marketing dataset (88\%) makes acceptance
rate a more informative primary metric than raw accuracy; the control plane's
keep rule (candidate $>$ current best) is applied to \texttt{val\_AUC}, which is
robust to imbalance. Both datasets are public on OpenML (IDs 31 and 1461), where
their licence is listed as ``Public'', and originate from the UCI Machine Learning
Repository.

\subsection{OpenML Running-Best Trajectories}
\label{sec:supp-fig-openml}

\begin{figure}[ht]
\centering
\includegraphics[width=\linewidth]{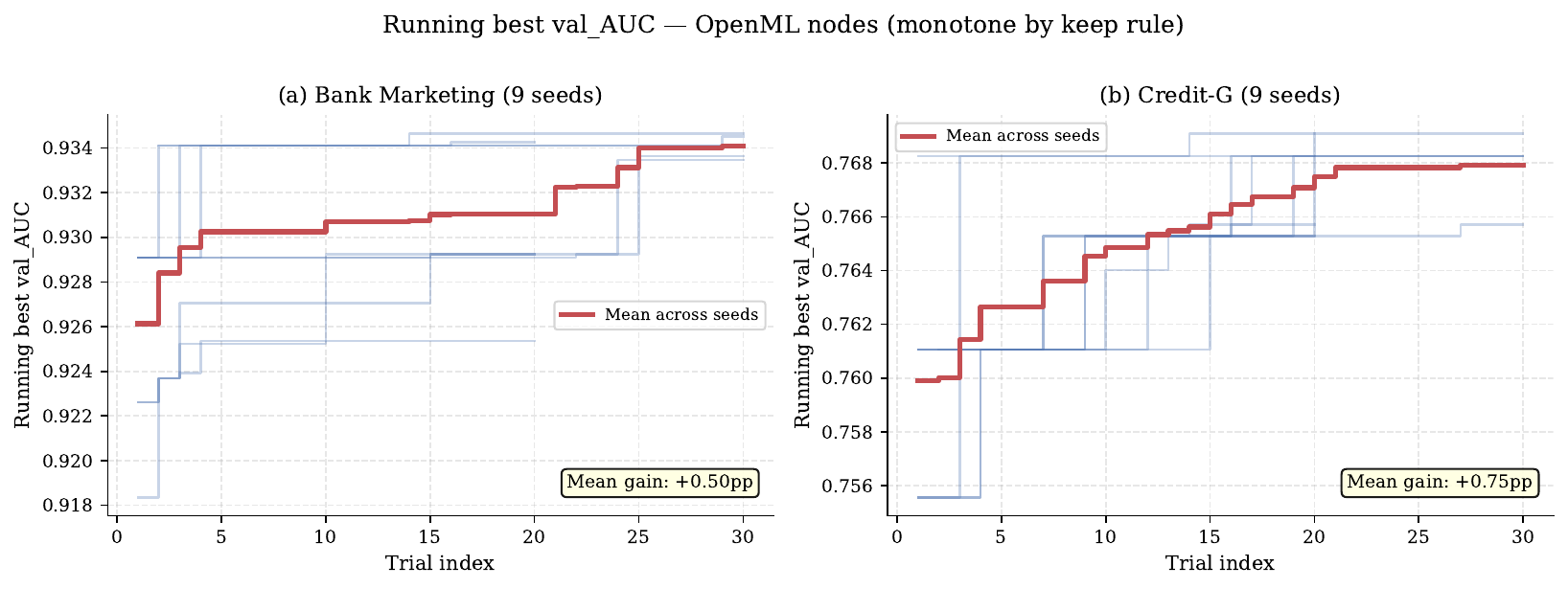}
\Description{OpenML Bank Marketing and Credit-G running-best validation AUC trajectories across seeds.}
\caption{OpenML running-best val\_AUC trajectories across all seeds. Monotone
non-decrease is guaranteed by the control-plane keep rule.
(a) Bank Marketing; (b) Credit-G.}
\label{fig:supp-openml}
\end{figure}

Figure~\ref{fig:supp-openml} shows running-best \texttt{val\_AUC} trajectories
for both OpenML nodes. The step function is monotonically non-decreasing because
the control plane accepts only valid improving trials --- the keep rule is enforced
structurally, not by the manager. Individual seeds (thin blue lines) and the
5-seed mean (red) show the variance compressed by the governed campaign discipline.

\section{Control-Plane Algorithm}
\label{sec:supp-algorithm}

\subsection{Trial Lifecycle}

Figure~\ref{alg:lifecycle} gives the pseudocode for one governed trial. The
control plane is the only writer to the append-only ledger; the manager never
transitions state directly.

\begin{figure}[ht]
\small
\renewcommand{\labelenumi}{\arabic{enumi}.}%
\begin{enumerate}
  \item Write a pending-guard file for this budget slot.
  \item Call \textbf{Manager.propose}(memory context, node spec) $\to$ proposal.
  \item \textbf{Scope-check:} if proposal touches any frozen path $\to$ reject.
  \item \textbf{No-op check:} if patch is empty or byte-identical $\to$ reject.
  \item \textbf{Precondition check:} validate proposal fields $\to$ reject if
        malformed.
  \item Call \textbf{Worker.execute}(proposal, node spec) $\to$ patch + raw log.
  \item If worker fails $\to$ record \texttt{failed\_invalid/runtime\_error} or
        \texttt{edit\_failed} and go to step 10.
  \item Parse task metric from raw log.
  \item If metric missing $\to$ \texttt{failed\_invalid/metric\_missing}. If
        metric $\leq$ current best $\to$ \texttt{discarded}. Else $\to$
        \texttt{kept}; update current best.
  \item Append one immutable ledger record (fields: proposal, patch ref, log ref,
        metric, decision, failure category, provenance IDs, node-state hash).
  \item Delete pending-guard file.
  \item Update append-only memory context from completed record.
\end{enumerate}
\Description{Pseudocode for one governed trial lifecycle from pending guard creation through manager proposal, validation, worker execution, decision, ledger append, and memory update.}
\caption{Trial lifecycle pseudocode (one budget slot).}
\label{alg:lifecycle}
\end{figure}

Steps 3--5 are pre-run guards: they prevent invalid edits from corrupting node
state and produce classified failure records before any training occurs.
Steps 7--9 are post-run classification: a valid but non-improving trial becomes
\texttt{discarded}, not \texttt{failed\_invalid}. The manager is not consulted at
any classification step.

\subsection{Control-Plane Trial Decision Timeline}
\label{sec:supp-fig-timeline}

\begin{figure}[ht]
\centering
\includegraphics[width=\linewidth]{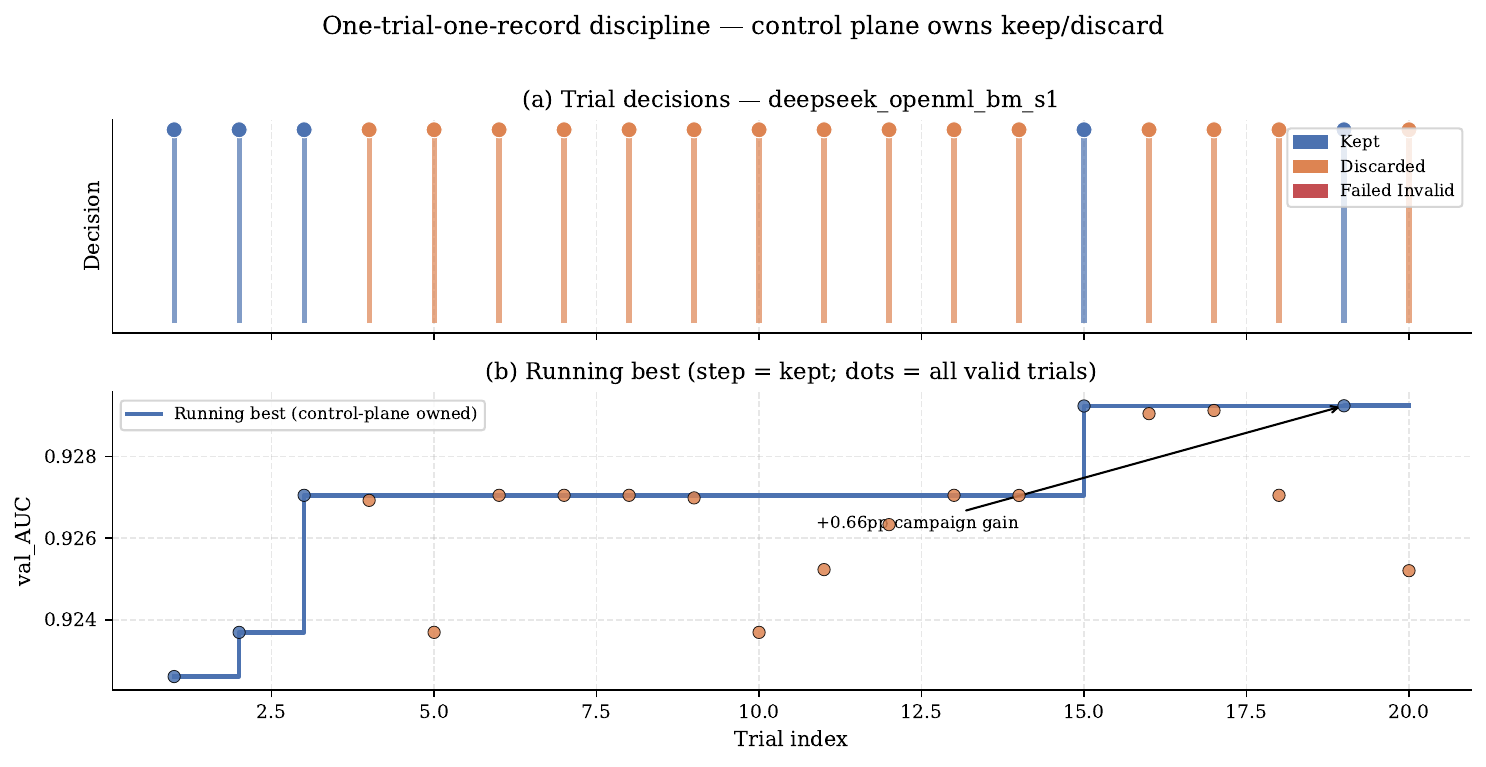}
\Description{Trial decision timeline for one OpenML Bank Marketing governed campaign with per-trial decisions and running-best validation AUC.}
\caption{Trial decision timeline for one governed campaign (OpenML Bank
Marketing, seed 1, 20 trials). Top: per-trial decision colour (kept / discarded /
failed\_invalid). Bottom: running best val\_AUC step function and per-trial metric
dots. The keep/discard decision is owned by the control plane, not the manager.}
\label{fig:supp-timeline}
\end{figure}

Figure~\ref{fig:supp-timeline} illustrates the one-trial-one-record discipline
on a representative campaign (OpenML Bank Marketing, seed~1). Every trial produces
exactly one terminal record. The running best (step function) advances only when
the control plane accepts a valid improving trial; it never decreases. This
monotone property is structural: the keep rule \texttt{candidate > current\_best}
is evaluated and applied exclusively by the control plane.

\subsection{Governed vs. Ungoverned Counterfactual Visual}
\label{sec:supp-counterfactual-visual}

\begin{figure*}[t]
\centering
\includegraphics[width=\linewidth]{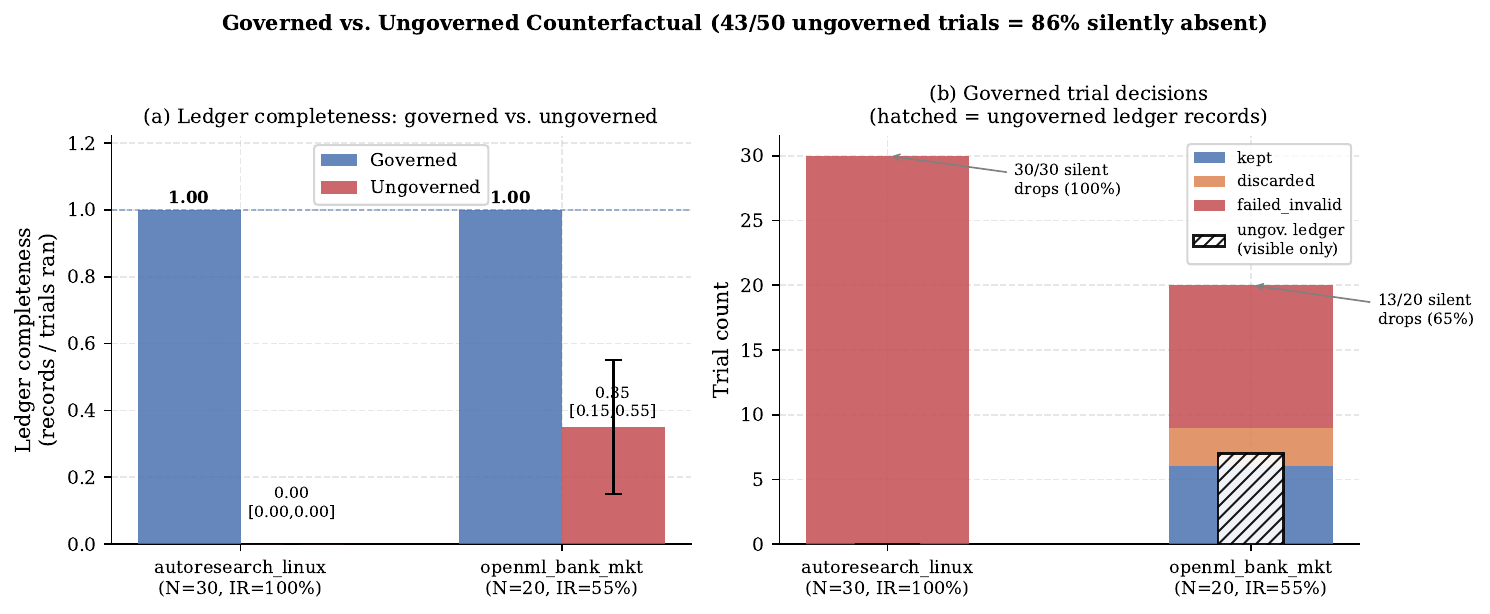}
\Description{Governed versus ungoverned ledger completeness with bootstrap confidence intervals and a governed trial decision timeline annotated with silent drops.}
\caption{Governed vs.\ ungoverned counterfactual (visual companion to
Table~\ref{tab:counterfactual}). Left: ledger completeness with 95\% bootstrap CIs; the
governed CI collapses to $[1.00,\,1.00]$ by structural invariant, while the
ungoverned CI lies strictly below 1.00 on both nodes. Right: governed trial
decision timeline with silent-drop positions annotated. Combined: 43\,/\,50
ungoverned trials, all of them failed trials, leave no ledger record; the two
campaigns were chosen to span failure rates.}
\label{fig:supp-counterfactual}
\end{figure*}

Figure~\ref{fig:supp-counterfactual} is the visual companion to the
counterfactual comparison of Table~\ref{tab:counterfactual}.

\subsection{ResNet L40S Per-Seed Panels}
\label{sec:supp-resnet-linux-panels}

\begin{figure}[t]
\centering
\includegraphics[width=\linewidth]{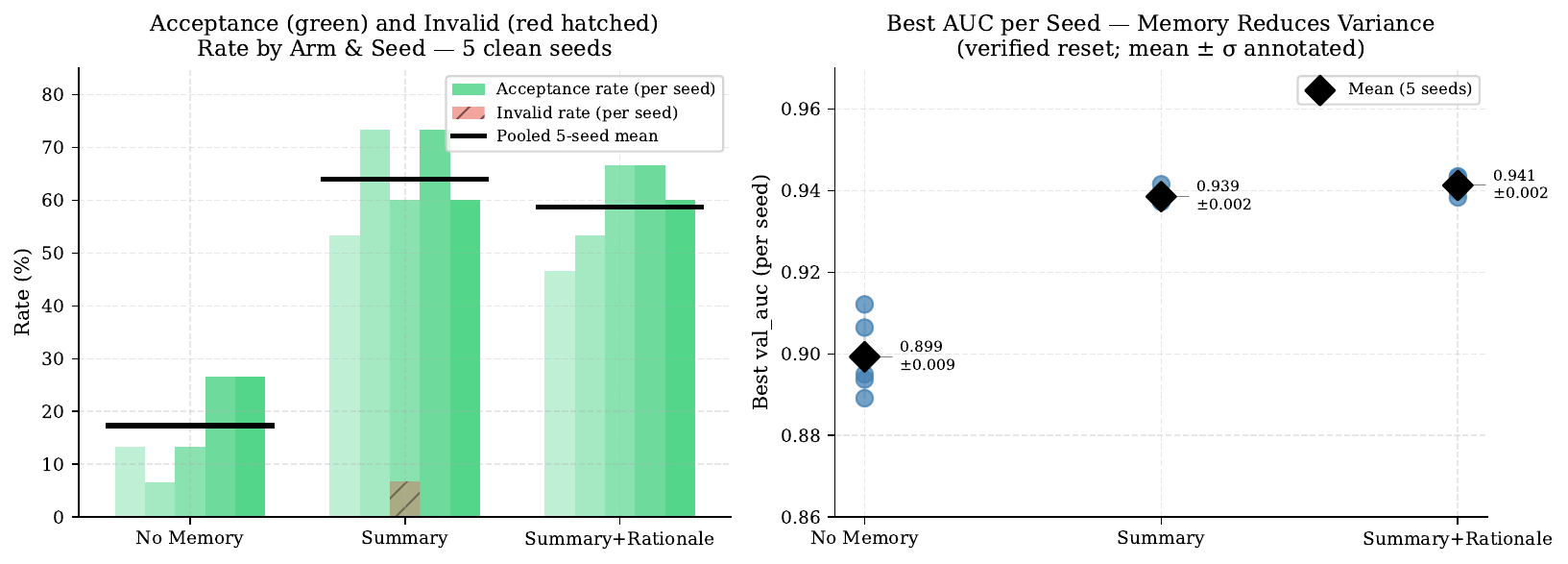}
\Description{ResNet L40S memory ablation across five clean seeds, showing per-seed acceptance and invalid rates plus per-seed best AUC.}
\caption{ResNet L40S memory ablation across five clean seeds with
verified reset. Left: per-seed acceptance (green) and invalid (red) rates.
Right: per-seed best AUC, showing $\sigma_{\text{none}} = 0.009$ vs.\
$\sigma_{\text{summary}} = \sigma_{\text{rationale}} = 0.002$ (a 4--5$\times$
smaller standard deviation; descriptive, five seeds per arm).}
\label{fig:supp-resnet-linux-panels}
\end{figure}

Figure~\ref{fig:supp-resnet-linux-panels} shows the per-seed acceptance and
invalid rates and the per-seed best AUC of the L40S replication.

\subsection{Decision Rule}
\label{sec:supp-decision-rule}

The decision function \texttt{decide\_trial()} is purely functional with no side
effects. It receives the candidate metric, the current-best metric, and a validity
flag, and returns exactly one terminal state:

\begin{figure}[ht]
\small
\begin{enumerate}\tightlist
  \item[\textbf{if}] validity flag is \textsc{false} (any pre-run or post-run guard
        fired) $\to$ return \texttt{failed\_invalid} with the guard's
        \texttt{FailureCategory}.
  \item[\textbf{elif}] $\texttt{candidate\_metric} > \texttt{current\_best\_metric}$
        $\to$ return \texttt{kept}; update current best to
        \texttt{candidate\_metric}.
  \item[\textbf{else}] $\to$ return \texttt{discarded} with
        \texttt{failure\_category = degraded\_metric}.
\end{enumerate}
The function never raises an exception and never reads or writes the ledger.
All side effects (ledger append, memory update, best-metric update) happen in the
control-plane campaign loop after \texttt{decide\_trial()} returns.
\Description{Pseudocode for the pure decision rule that maps validity and candidate metrics to failed-invalid, kept, or discarded.}
\caption{Decision rule pseudocode (\texttt{control\_plane/decision.py}).}
\label{fig:decision-rule}
\end{figure}

\subsection{Memory Context Algorithm}
\label{sec:supp-memory-algorithm}

The memory context is built by \texttt{build\_memory\_context()} and injected into
the manager at each trial. Three modes produce qualitatively different inputs:

\begin{figure}[ht]
\small
\textbf{Mode: none} \\
Return a context containing only the NodeSpec parameters (editable symbols and
their current values). No prior trial history is included. The manager has no
information about past proposals or outcomes.

\smallskip
\textbf{Mode: summary} \\
Append one line per completed trial in chronological order:
\begin{lstlisting}
Trial {i}: {decision} - {proposal_summary}
  metric: {parsed_metrics}
\end{lstlisting}
The manager receives the full history of proposals and outcomes, but with no
explanation of \emph{why} each trial succeeded or failed.

\smallskip
\textbf{Mode: rationale (summary + rationale)} \\
Same as summary, plus per-trial failure rationale lines:
\begin{lstlisting}
Trial {i}: {decision} - {proposal_summary}
  metric: {parsed_metrics}
  rationale: {proposal_rationale}
  [AVOID: repeated {target} change - prior rejection]
\end{lstlisting}
The \texttt{[AVOID:~...]} warning is appended whenever the repeated-bad detector
flags the current trial as semantically repeating a prior failed edit (see
Section~\ref{sec:supp-rbr}). Rationale tokens can be truncated to a configurable
\texttt{rationale\_max\_tokens} limit (50 tokens in the verbosity ablation).
\Description{Pseudocode-style description of none, summary, and rationale memory context construction.}
\caption{Memory context algorithm (\texttt{memory/summarizer.py}).}
\label{fig:memory-algorithm}
\end{figure}

\subsection{Repeated-Bad Rate Detection}
\label{sec:supp-rbr}

The repeated-bad rate (RBR) is a governance metric that measures how often the
manager repeats the same edit target and mechanism as a prior rejected trial
without providing a new corrective rationale.

\textbf{Definition.} Let $P_i = (t_i, d_i)$ be the edit target and direction of
proposal $i$ (e.g., target = \texttt{LEARNING\_RATE}, direction = \emph{increase}).
Proposal $i$ is \emph{repeated-bad} if:
\begin{enumerate}\tightlist
  \item A prior proposal $P_j$, $j < i$, had the same target $t_j = t_i$ and
        direction $d_j = d_i$; and
  \item $P_j$ received decision \texttt{discarded} or \texttt{failed\_invalid}; and
  \item Proposal $i$ does not include a rationale that references the prior
        rejection.
\end{enumerate}
$\mathrm{RBR} = |\{i : P_i \text{ is repeated-bad}\}| / N$
where $N$ is the campaign budget.

\textbf{Implementation.} The detector in \texttt{memory/\allowbreak similarity.py}
uses Jaccard similarity on unigram token sets of proposal summaries to identify
near-duplicate proposals, then checks whether the candidate's summary token set
$S_i$ overlaps with any prior rejected proposal $S_j$ above a threshold
$\theta = 0.6$:
\[
  \mathrm{Jaccard}(S_i, S_j) = \frac{|S_i \cap S_j|}{|S_i \cup S_j|} \geq \theta
\]
Parameter direction is extracted by regex from the proposal rationale field
(keywords: \emph{increase}, \emph{decrease}, \emph{reduce}, \emph{raise},
\emph{lower}). Proposals that lack a detectable direction are not flagged.

\subsection{Manager Prompt Template}
\label{sec:supp-prompt-template}

The LangGraph manager (\texttt{manager/\allowbreak langgraph\_manager.py}) builds a
single-turn prompt from five components at each trial. Below is the template
structure; bracketed tokens are filled at runtime.

\begin{lstlisting}[basicstyle=\scriptsize\ttfamily,breaklines=true,
  breakatwhitespace=true,columns=flexible]
You are a research manager proposing one bounded
hyperparameter change.

Node: [node_spec.name]
Description: [node_spec.description]
Metric to optimize: [metric_name] (direction: [direction])
Files you may instruct the worker to edit: [editable_paths]
Budget index (trial number): [budget_index]
Current best [metric_name]: [current_best_metric]

Editable values currently in [editable_file]:
  PARAM_A = current_value_A
  PARAM_B = current_value_B
  ...

# --- if memory_context is non-empty: ---
Prior trial memory:
[memory_context.context_text]

AVOIDANCE RULE (mandatory): Study the trial history
above carefully. Identify every hyperparameter that
was already tried in the same direction (e.g.
increasing BATCH_SIZE, reducing LEARNING_RATE) and
produced no improvement or a failure. You MUST NOT
propose the same parameter change in the same
direction again. Choose a DIFFERENT hyperparameter
or a DIFFERENT direction. If every direction for a
parameter has been explored, switch to a different
parameter entirely.

# --- if memory_context is empty: ---
No prior trial memory available.

Propose exactly one bounded experiment. Respond with
ONLY a JSON object (no markdown, no extra text):
  "summary"   -- short slug e.g. "reduce-lr-2e-4"
  "rationale" -- one sentence on why this might help
  "objective" -- full worker instruction:
                 "In FILE, change PARAM from X to Y.
                  Edit only FILE."
  "param"     -- exact editable value name
  "old_value" -- current value as listed above
  "new_value" -- proposed replacement value

Constraints:
  - edit only: [editable_paths]
  - change exactly one listed value
  - param must be one of the editable values above
  - do not run the experiment; the harness runs it
\end{lstlisting}

The AVOIDANCE RULE is injected only when prior trial memory is non-empty
(i.e., after the first trial). It operates on the memory context produced by the
mode selected for that campaign arm. In the \texttt{none} mode the rule is absent;
in the \texttt{summary} and \texttt{rationale} modes it is present and the manager
is expected to honour it. The RBR metric (Section~\ref{sec:supp-rbr}) measures
empirical compliance.

\subsection{Scope Validation}

The control plane's scope validator
checks the file set touched by the patch against the NodeSpec's
\texttt{editable\allowbreak\_paths} whitelist and \texttt{frozen\allowbreak\_paths} blocklist. A patch
is rejected if it touches any file outside the whitelist or any file in the
blocklist, regardless of the nature of the change. This fires before the worker
executes training, so frozen files (dataset pipelines, split logic, evaluation
harness) are never modified.

\subsection{Crash-Recovery Path}

If the process crashes mid-trial (e.g.\ OOM kill, host reboot), the pending-guard
file persists on disk. At the start of the next campaign run,
\texttt{PendingTrialError} is raised. The operator runs:

\begin{lstlisting}
python3 scripts/recover_pending.py --list
python3 scripts/recover_pending.py --inspect \
    experiments/ledgers/<id>_pending.json
python3 scripts/recover_pending.py --mark-failed <id> \
    --reason "OOM during training"
\end{lstlisting}

This appends a \texttt{failed\_invalid} record for the stale trial before any new
trial begins, preserving the invariant that every opened budget slot produces
exactly one terminal ledger record. The guard file is deleted only after the
failure record is written.

\section{NodeSpec Contract Specification}
\label{sec:supp-nodespec}

\subsection{Schema}

A NodeSpec is a JSON-serialisable YAML file under \texttt{configs/\allowbreak nodes/}.
The control plane reads it once at campaign start and treats it as immutable for
the duration of the campaign. Table~\ref{tab:supp-nodespec-schema} lists the required
and optional fields.

\begin{table}[ht]
\centering
\footnotesize
\caption{NodeSpec YAML field schema. All fields are required unless marked
optional.}
\label{tab:supp-nodespec-schema}
\begin{tabular}{@{}p{0.30\linewidth}p{0.14\linewidth}p{0.48\linewidth}@{}}
\toprule
Field & Type & Meaning \\
\midrule
\texttt{name} & string & Unique node identifier used as campaign ledger key. \\
\texttt{description} & string & Human-readable node purpose. \\
\texttt{editable\_paths} & list & Files the worker may edit; scope validator
  enforces this whitelist. \\
\texttt{frozen\_paths} & list & Files and directories the worker must never
  touch; scope validator enforces this blocklist. \\
\texttt{run\_command} & string & Shell command executed by the worker to train
  and produce output. \\
\texttt{metric\_name} & string & Key expected in parsed metric output
  (e.g.\ \texttt{val\_auc}). \\
\texttt{metric\_direction} & string & \texttt{maximize} or \texttt{minimize};
  determines the keep/discard rule. \\
\texttt{metric\_parser} & string & Python import path to the parser function
  (\texttt{module:function}). \\
\texttt{acceptance\_rule} & string & Expression evaluated by decision module
  (e.g.\ \texttt{candidate\_metric > current\_best\_metric}). \\
\texttt{default\_budget} & object & Default \texttt{trials} and
  \texttt{max\_wall\_clock\_hours} (both optional at invocation time). \\
\texttt{setup\_command} & string & Optional pre-campaign setup (e.g.\
  \texttt{uv sync}). \\
\texttt{editable\_symbols} & list & Optional: specific named symbols inside
  editable files (used by LocalWorker). \\
\texttt{failure\_categories} & list & Exhaustive list of valid
  \texttt{FailureCategory} values for this node. \\
\bottomrule
\end{tabular}
\end{table}

\subsection{Annotated Example: \texttt{resnet\_trigger}}

This is the primary scientific node. The worker may only edit \texttt{train.py};
all other files including datasets, split logic, and architecture are frozen.

\begin{lstlisting}[language={}]
name: resnet_trigger
description: >
  Near-threshold detector waveform binary-classification benchmark
  with frozen H5 signal/noise traces and a single editable 1D ResNet
  training entrypoint.
editable_paths:
  - train.py
frozen_paths:
  - prepare.py
  - resnet_1d.py
  - signal_vacuum_sum_crop_4000x8000.h5
  - noise_traces_4000x8000.h5
  - artifacts/
  - data/
run_command: >
  ~/anaconda3/bin/python3 train.py > run.log 2>&1
metric_name: val_auc
metric_direction: maximize
metric_parser: >
  autoresearch.nodes.resnet_trigger.metric_parser:parse_val_auc
acceptance_rule: candidate_metric > current_best_metric
default_budget:
  trials: 50
  max_wall_clock_hours: 12
\end{lstlisting}

The metric parser converts \texttt{val\_bpb} to \texttt{val\_auc} via
$1 - \texttt{val\_bpb}$ when both values are present, preferring the
\texttt{val\_bpb} reading. This indirection allows the same governance
infrastructure to handle both maximise- and minimise-direction nodes.

\subsection{Annotated Example: \texttt{lr\_synthetic}}

This is the dependency-free transfer node. The worker edits \texttt{train.py}
but is further restricted to four named symbols; the data-generation constants
are frozen by scope enforcement and by the \texttt{editable\_symbols} hint used
by LocalWorker.

\begin{lstlisting}[language={}]
name: lr_synthetic
description: >
  Synthetic binary-classification node (NumPy logistic regression).
  Dependency-free governance-protocol validation.
editable_paths:
  - train.py
editable_symbols:
  - LEARNING_RATE
  - REGULARIZATION
  - N_EPOCHS
  - BATCH_SIZE
frozen_paths:
  - artifacts/
run_command: python3 train.py
metric_name: val_score
metric_direction: maximize
metric_parser: >
  autoresearch.nodes.lr_synthetic
  .metric_parser:parse_val_score
acceptance_rule: candidate_metric > current_best_metric
default_budget:
  trials: 10
  max_wall_clock_hours: 1
\end{lstlisting}

Adding a new node requires only a YAML file in \texttt{configs/\allowbreak nodes/}
and a registration entry in
\texttt{src/autoresearch/\allowbreak nodes/\allowbreak registry.py}.
The control-plane contract (lifecycle, guards, ledger format) is identical.

\section{Failure Taxonomy with Concrete Examples}
\label{sec:supp-failure-taxonomy}

The control plane assigns a \texttt{FailureCategory} label at the terminal
transition that writes the ledger record. The taxonomy is exhaustive for the
current NodeSpec contract: every \texttt{failed\_invalid} outcome maps to exactly
one of the categories below. Table~\ref{tab:supp-failure-taxonomy} lists each
category, its detection point, and a representative ledger excerpt.

\begin{table*}[t]
\centering
\footnotesize
\caption{Failure taxonomy: detection point, guard type, and representative ledger
evidence. All examples are drawn from real campaign ledgers in the released
evidence; fields are formatted for readability.}
\label{tab:supp-failure-taxonomy}
\begin{tabular}{@{}p{0.18\linewidth}p{0.12\linewidth}p{0.14\linewidth}p{0.49\linewidth}@{}}
\toprule
Category & Fires at & Guard & Representative ledger evidence \\
\midrule
\texttt{invalid\_edit\_scope} & Pre-run &
  Scope validator &
  Campaign \texttt{kdd\_stress\_scope}, trial-001: proposal attempts to edit
  \texttt{resnet\_1d.py} (frozen); scope guard rejects; decision is
  \texttt{failed\_invalid}; patch is never applied; node state is unchanged. \\
\texttt{no\_op\_patch} & Pre-run &
  No-op guard &
  Campaign \texttt{kdd\_stress\_noop}, trial-001: worker produces a patch that
  is byte-identical to the current \texttt{train.py}; no-op guard rejects;
  decision is \texttt{failed\_invalid}; training is never executed. \\
\texttt{proposal\_precondition\_\allowbreak failed} & Pre-run &
  Precondition check &
  Campaign \texttt{deepseek\allowbreak\_autoresearch\allowbreak\_linux\allowbreak\_append\allowbreak\_only\allowbreak\_summary\allowbreak\_s2},
  trials 28--30: manager produces malformed proposals after repeated
  \texttt{runtime\allowbreak\_error} failures; precondition guard fires before
  worker call. \\
\texttt{runtime\_error} & Post-run &
  Metric/exit check &
  Campaigns \texttt{deepseek\allowbreak\_autoresearch\allowbreak\_linux\allowbreak\_none\allowbreak\_s1}--\texttt{s4},
  all 120 trials: the worker exits non-zero (cause undetermined; see
  Section~\ref{autoresearch-linux}); provenance IDs complete in all 120
  records. \\
\texttt{invalid\_config} & Post-run &
  Node validator &
  OpenML nodes reject hyperparameter values outside their bounded configuration
  domains, e.g.\ excessive tree depth or estimator counts.
  The run emits a machine-readable \texttt{invalid\_config} label, which the
  control plane records as \texttt{failed\_invalid}. \\
\texttt{metric\_missing} & Post-run &
  Metric parser &
  Occurs when a run completes (exit zero) but the log file contains no parseable
  metric line, e.g.\ a training script that writes to \texttt{stderr} only or
  silently skips the validation epoch. Decision is \texttt{failed\_invalid}; the
  raw log ref is retained for inspection. \\
\texttt{edit\_failed} & Post-run &
  Worker path &
  ClawWorker only: the coding agent produces a patch file but fails to apply it
  cleanly (merge conflict, syntax error introduced by the patch). This is a
  worker-path artifact; it does not appear in LocalWorker or LangGraph
  deterministic-patch campaigns. \\
\midrule
\texttt{degraded\_metric} & Post-run &
  Decision rule &
  \textit{Not} \texttt{failed\_invalid}: valid run, metric parses, but does not
  beat current best. Decision is \texttt{discarded}. E.g.\ campaign
  \texttt{kdd\_resnet\_scientific\_20}, trial-002: patch runs successfully,
  \texttt{val\_auc} = 0.7738 vs.\ current best 0.7747; trial is discarded with
  full artifact retention. \\
\bottomrule
\end{tabular}
\end{table*}

\subsection{Concrete Ledger Records}
\label{sec:supp-ledger-records}

{\sloppy The following records are drawn verbatim from the campaign ledgers (provenance
hash suffixes truncated for readability; full records are in
\texttt{experiments/\allowbreak ledgers/}). Each record shows the seven fields that
identify a trial's terminal outcome. The \texttt{metric\_missing} category has not
been observed in any real campaign to date; all zero-metric outcomes in the
released ledgers are classified as \texttt{runtime\_error} (the worker exits
non-zero before the metric parser runs).\par}

\noindent\textbf{Kept} (campaign \texttt{kdd\_resnet\_scientific\_20}, trial-001):
\begin{lstlisting}[basicstyle=\scriptsize\ttfamily,breaklines=true,columns=flexible]
{"trial_id": "kdd_resnet_scientific_20-trial-001",
 "campaign_id": "kdd_resnet_scientific_20",
 "budget_index": 1,
 "decision": "kept",
 "failure_category": null,
 "proposal_summary": "lower-weight-decay",
 "parsed_metrics": {"val_auc": 0.774711}}
\end{lstlisting}

{\sloppy\noindent\textbf{Discarded} (campaign
\texttt{kdd\_resnet\_scientific\_20}, trial-002):\par}
\begin{lstlisting}[basicstyle=\scriptsize\ttfamily,breaklines=true,columns=flexible]
{"trial_id": "kdd_resnet_scientific_20-trial-002",
 "campaign_id": "kdd_resnet_scientific_20",
 "budget_index": 2,
 "decision": "discarded",
 "failure_category": "degraded_metric",
 "proposal_summary": "small-dropout",
 "parsed_metrics": {"val_auc": 0.773778}}
\end{lstlisting}

{\sloppy\noindent\textbf{invalid\_edit\_scope} (campaign
\texttt{kdd\_stress\_scope}, trial-001):\par}
\begin{lstlisting}[basicstyle=\scriptsize\ttfamily,breaklines=true,columns=flexible]
{"trial_id": "kdd_stress_scope-trial-001",
 "campaign_id": "kdd_stress_scope",
 "budget_index": 1,
 "decision": "failed_invalid",
 "failure_category": "invalid_edit_scope",
 "proposal_summary": "lower-learning-rate",
 "parsed_metrics": null}
\end{lstlisting}
The scope guard fires before the worker executes; node state is never modified
and training never runs.

\noindent\textbf{no\_op\_patch} (campaign \texttt{kdd\_stress\_noop}, trial-001):
\begin{lstlisting}[basicstyle=\scriptsize\ttfamily,breaklines=true,columns=flexible]
{"trial_id": "kdd_stress_noop-trial-001",
 "campaign_id": "kdd_stress_noop",
 "budget_index": 1,
 "decision": "failed_invalid",
 "failure_category": "no_op_patch",
 "proposal_summary": "lower-weight-decay",
 "parsed_metrics": null}
\end{lstlisting}

{\sloppy\noindent\textbf{proposal\_precondition\_failed}\\
(campaign \texttt{deepseek\_autoresearch\_linux\_\allowbreak{}append\_only\_summary\_s1},
trial-012):\par}
\begin{lstlisting}[basicstyle=\scriptsize\ttfamily,breaklines=true,columns=flexible]
{"trial_id":
   "deepseek_autoresearch_linux_append_only_summary_s1-trial-012",
 "campaign_id":
   "deepseek_autoresearch_linux_append_only_summary_s1",
 "budget_index": 12,
 "decision": "failed_invalid",
 "failure_category": "proposal_precondition_failed",
 "proposal_summary": "reduce-total-batch-size-32768",
 "parsed_metrics": null}
\end{lstlisting}
The precondition check fires when the manager produces a proposal whose field
values fail validation (e.g.\ \texttt{new\_value} is identical to
\texttt{old\_value}, or a required field is absent). No worker call is made.

{\sloppy\noindent\textbf{runtime\_error}\\
(campaign \texttt{deepseek\_autoresearch\_linux\_none\_s1}, trial-001):\par}
\begin{lstlisting}[basicstyle=\scriptsize\ttfamily,breaklines=true,columns=flexible]
{"trial_id":
   "deepseek_autoresearch_linux_none_s1-trial-001",
 "campaign_id": "deepseek_autoresearch_linux_none_s1",
 "budget_index": 1,
 "decision": "failed_invalid",
 "failure_category": "runtime_error",
 "proposal_summary": "reduce-embedding-lr-0.01",
 "parsed_metrics": null}
\end{lstlisting}
All 120 trials in the \texttt{none}-memory arm of the autoresearch-linux node share
this pattern: the worker exits non-zero and the control plane records
\texttt{runtime\_error}. The cause is undetermined because the start state of
these campaigns is unverified (Section~\ref{autoresearch-linux}).

{\sloppy\noindent\textbf{invalid\_config}\\
(campaign \texttt{deepseek\_openml\_bm\_s2}, trial-005):\par}
\begin{lstlisting}[basicstyle=\scriptsize\ttfamily,breaklines=true,columns=flexible]
{"trial_id":
   "deepseek_openml_bm_s2-trial-005",
 "campaign_id": "deepseek_openml_bm_s2",
 "budget_index": 5,
 "decision": "failed_invalid",
 "failure_category": "invalid_config",
 "proposal_summary": "increase-n-estimators-700",
 "parsed_metrics": null}
\end{lstlisting}
The \texttt{invalid\_config} category covers node-level constraint violations
where the edited file is in scope but the proposed value is outside the
NodeSpec's bounded domain. It appears in the primary OpenML evidence set and is
therefore included in the taxonomy of Section~\ref{failure-taxonomy}.

\subsection{Taxonomy Reliability Probe}

{\sloppy
To check that failure categories are structurally assignable from observable
ledger evidence, we applied an independent rule-based relabeler to a
stratified sample of 50 \texttt{failed\_invalid} records spanning the observed
validation categories. The relabeler uses only scope-validation flags, patch presence,
training-initiation signals, and execution status --- never the
\texttt{failure\_category} field itself. Cohen's~$\kappa$ between the
control-plane assignments and the structural relabeler is
$\kappa = 0.830$ (95\,\% CI $[0.703, 0.958]$), Landis--Koch ``almost
perfect'' agreement. This is a structural agreement check, not a human
inter-rater study. Perfect agreement ($\kappa=1.0$) is achieved for
\texttt{invalid\_config}, \texttt{invalid\_edit\_scope}, and
\texttt{proposal\-\_precondition\-\_failed}; the six discordant cases are
\texttt{runtime\_error} and \texttt{no\_op\_patch} edge cases where the
distinguishing evidence (training seed, no-op message) is absent from the
JSONL record but present in the referenced artifact files. Full human
inter-rater validation remains future work.\par}

\subsection{Distinguishing Failed from Discarded}

Key design choice: valid non-improving trials are \texttt{discarded},
not \texttt{failed\allowbreak\_invalid}. This matters for audit:
\begin{itemize}
  \item \texttt{failed\allowbreak\_invalid}: a process problem (scope violation,
        coding failure, missing output) the operator should inspect.
  \item \texttt{discarded} trials indicate a valid experiment that did not improve
        the metric --- a normal outcome in exploration campaigns with high
        discard rates.
\end{itemize}
Treating discard as failure would inflate the invalid rate and make it impossible
to distinguish bad proposals from bad outcomes.

\subsection{New Category Protocol}
\label{sec:supp-new-category}

\begin{sloppypar}
To add a new failure category:
\begin{enumerate}\tightlist
  \item Add the enum label to \texttt{Failure\-Category}
        in \texttt{memory/\allowbreak schemas.py}.
  \item Add detection logic in \texttt{control\_\allowbreak plane/\allowbreak campaign.py}
        at the appropriate guard or post-run classification point.
  \item Add the name to \texttt{failure\allowbreak\_categories} in the NodeSpec
        YAML(s).
  \item Add a test in \texttt{tests/\allowbreak test\_trial\_schema.py}.
\end{enumerate}
The new category must map to \texttt{kept}, \texttt{discarded}, or
\texttt{failed\allowbreak\_invalid} before the ledger append.
\end{sloppypar}

\end{document}